\documentclass[conference, compsoc]{IEEEtran}
\IEEEoverridecommandlockouts
\usepackage{titletoc}
\usepackage{cite}
\usepackage{amsmath,amssymb,amsfonts}
\usepackage{graphicx}
\usepackage{textcomp}
\usepackage[table]{xcolor}
\usepackage{blindtext}
\usepackage{floatrow}
\usepackage{url}
\usepackage[utf8]{inputenc}
\usepackage{soul}
\usepackage{hyperref}
\hypersetup{
  colorlinks   = true, 
  urlcolor     = black, 
  linkcolor    = black, 
  citecolor   = black 
}
\usepackage{newfloat}
\usepackage{listings}
\usepackage{amsfonts}
\usepackage{enumitem}
\usepackage{algorithm}
\usepackage[noend]{algpseudocode}
\usepackage{mathtools}
\usepackage{makecell}
\usepackage{multirow}
\usepackage{subcaption}
\usepackage{dsfont}
\usepackage{booktabs}
\usepackage[switch]{lineno}

\DeclareMathOperator*{\maximize}{maximize}

\DeclareMathOperator*{\argmax}{arg\,max}

\newcommand\tunedown[1]{{\color{black}{#1}}}

\DeclareCaptionStyle{ruled}{labelfont=normalfont,labelsep=colon,strut=off} 
\floatstyle{ruled}
\newfloat{listing}{tb}{lst}{}
\floatname{listing}{Listing}

\def\BibTeX{{\rm B\kern-.05em{\sc i\kern-.025em b}\kern-.08em
    T\kern-.1667em\lower.7ex\hbox{E}\kern-.125emX}}

\begin{document}

\title{MM-BD: Post-Training Detection of Backdoor Attacks with Arbitrary Backdoor Pattern Types Using a Maximum Margin Statistic}


\author{
\IEEEauthorblockN{Hang Wang\textsuperscript{\rm *}, Zhen Xiang,\textsuperscript{\rm *} David J. Miller\textsuperscript, George Kesidis\textsuperscript}
\IEEEauthorblockA{
\textit{Anomalee Inc.}, State College, PA, USA 
\textit{Pennsylvania State University}, University Park, PA, USA
}
}
\IEEEaftertitletext{\vspace{-1\baselineskip}}

\maketitle

\def\thefootnote{*}\footnotetext{Equal contribution}\def\thefootnote{\arabic{footnote}}
\begin{abstract}
Backdoor attacks are an important type of adversarial threat against deep neural network classifiers, wherein test samples from one or more source classes will be (mis)classified to the attacker's target class when a backdoor pattern is embedded.
In this paper, we focus on the {\it post-training} backdoor defense scenario commonly considered in the literature, where the defender aims to detect whether a trained classifier was backdoor-attacked without any access to the training set.
\tunedown{Many post-training detectors are designed to detect attacks that use either one or a few specific backdoor embedding functions (e.g., patch-replacement or additive attacks)}.
These detectors may fail when the backdoor embedding function used by the attacker (unknown to the defender) is different from the backdoor embedding function assumed by the defender.
In contrast, we propose a post-training defense that detects backdoor attacks with {\it arbitrary} types of backdoor embeddings, without making any assumptions about the backdoor embedding type.
Our detector leverages the influence of the backdoor attack, {\it independent of the backdoor embedding mechanism}, on the landscape of the classifier's outputs prior to the softmax layer.
For each class, a {\it maximum margin} statistic is estimated.  Detection inference is then performed by applying an {\it unsupervised} anomaly detector to these statistics.
\tunedown{Thus, our detector does not need any legitimate clean samples, and can efficiently detect backdoor attacks with {\it arbitrary} numbers of source classes.}
These advantages over several state-of-the-art methods are demonstrated on four datasets, for three different types of backdoor patterns, and for a variety of attack configurations.
Finally, we propose a novel, general approach for backdoor mitigation once a detection is made. 
The mitigation approach was the runner-up at the first IEEE Trojan Removal Competition. \href{https://github.com/wanghangpsu/MM-BD}{The code is online available.} 
\end{abstract}


\vspace{-0.1in}
\section{Introduction}\label{sec:introduction}
\vspace{-0.1in}

Although deep neural networks (DNNs) are successful in many research areas, they are vulnerable to 
attacks 
\cite{Szegedy_seminal}.
A backdoor attack or Trojan is an important type of 
attack under which a DNN classifier will predict to the attacker's target class when a test sample from one or more source classes is embedded with the attacker's backdoor pattern \cite{BadNet, Trojan, BAsurvey}. A backdoor attack is typically launched by poisoning the classifier's training set with samples originally from the source classes, embedded with the same backdoor pattern that will be used during inference, and labeled to the target class \cite{Targeted}. Since successful backdoor attacks do not degrade the classifier's accuracy on clean test samples, they cannot be easily detected, e.g., using validation set accuracy \cite{Roni}. 

Defenses against backdoor attacks are sometimes deployed during the classifier's training stage \cite{SS, AC, shan2022poison, Differential_Privacy, huang2022backdoor, trim_loss, Hong20GS, doesntkill}, but this is also often infeasible (e.g., considering proprietary and legacy systems). Here, we consider a practical {\it post-training} scenario, where the defender is, e.g., a downstream user of a downloaded pre-trained classifier who aims to detect whether the classifier has been attacked (e.g. Model Zoo that provides pre-trained models \cite{ModelZoo}.),
without access to
the classifier's training set. The defender also has no access to any unattacked classifiers for reference (e.g. for setting a detection threshold).

\tunedown{Many post-training defenses assume the defender independently possesses a small set of clean, legitimate samples from every class.} These samples may be used: i) to reverse-engineer putative backdoor patterns, which are the basis for anomaly detection 
\cite{NC,Tabor,Post-TNNLS,ABS, TwoClass, Shen2021BackdoorSF,DataLimited}; or ii) to train shadow neural networks with and without (known) backdoor attacks – based upon which a binary ``meta-classifier'' is trained to predict whether the classifier under inspection was backdoor-attacked \cite{meta_sup, META, TrojAI}. However, these methods assume the mechanism used by the attacker for embedding a backdoor pattern is known. For reverse-engineering-based defenses, the embedding function of the backdoor pattern is explicitly involved in the reverse-engineering problem, e.g., \cite{NC}; thus, these methods may not be able to effectively detect backdoor attacks with pattern types different from that assumed for the reverse-engineering. For the meta-classification approach, a pool of backdoor patterns is assumed for training shadow neural networks with backdoor attacks \cite{META}; these methods may not generalize well to backdoor patterns not seen in the training pool.

Here, we propose a {\bf M}aximum-{\bf M}argin-based {\bf B}ackdoor {\bf D}etection method (MM-BD) \tunedown{which
does not make any assumptions about the backdoor pattern type used by the attacker. }
Our detection method is based on 
a novel approach to capture 
the atypicality of the landscape of the classifier's outputs (before the softmax) induced by backdoor attacks. In particular, we propose a novel {\it maximum margin} (MM) detection statistic.  This statistic is
obtained for each class by solving a margin maximization problem starting from multiple random input patterns, with the solution with largest margin then chosen.
We will show that use of these statistics is
discriminative between the backdoor target class and non-backdoor classes, irrespective of the type of backdoor pattern/embedding function used.
{\it No clean samples} are required for MM-based detection. A fully {\it unsupervised} anomaly detector is applied to the maximum margin statistics to perform the detection inference. Notably, the design of our method allows it to detect backdoor attacks with an {\it arbitrary} number of source classes, and also with better {\it computational efficiency} than many existing methods. 

We also propose a  {\bf M}aximum-{\bf M}argin-based {\bf B}ackdoor {\bf M}itigation method (MM-BM) that does require use of a few clean samples. This method suppresses the maximum possible neuron activations using a set of optimized upper bounds (one per neuron), without 
reducing the classifier's accuracy on clean samples. Unlike existing methods, we do not modify the DNN architecture or any trained parameters. 

Our contributions are summarized as follows:
\begin{itemize}[leftmargin=*]
    \item We reveal that the maximum margin (MM) can be used as a signature of the attack (for attacks with a sufficiently high attack success rate) in the landscape of the victim classifier's output function, irrespective of the backdoor pattern type. Based on this, we propose MM-BD, a post-training backdoor detection method without any assumptions about the backdoor pattern type used by the attacker. 
    \item \tunedown{MM-BD does not require any clean samples for detection.} Moreover, 
    as shown experimentally, it accurately, efficiently detects backdoor attacks 
    irrespective of the number of source classes used by the attacker.
    \item We show superior performance of MM-BD compared with many recent post-training detectors for standard attack settings considered by these works.  The experiments involve four datasets, three different types of backdoor patterns, and various DNN architectures.
    \item We evaluate MM-BD against many emerging backdoor attacks not considered by previous detectors. These attacks involve two additional types of backdoor patterns and six attack settings. Most of these attacks assume the attacker has strong control over the training process; by contrast, very few assumptions are made for MM-BD. 
    \item We propose a novel backdoor mitigation approach (MM-BM) by applying an optimized upper bound to each neuron activation. This approach does not modify the DNN architecture or any trainable parameters.
\end{itemize}

\vspace{-0.05in}
\section{Background}\label{sec:background}
\vspace{-0.05in}

This section provides background on backdoor attacks and defenses. We start with high-level descriptions of emerging threats to machine learning (ML) systems in Sec. \ref{subsec:adv_attacks}. In Sec. \ref{subsec:backdoor_attack}, we introduce the classical backdoor attack, including its goals and launching strategies. In Sec. \ref{subsec:backdoor_defenses} and Sec. \ref{subsec:backdoor_attack_advanced}, we give a taxonomy of backdoor defenses and advanced backdoor attacks, respectively.

\vspace{-0.1in}
\subsection{Threats to Machine Learning Systems}\label{subsec:adv_attacks}
\vspace{-0.05in}

Machine learning (ML) systems, especially deep neural networks, are starting to be adopted in various
safety-critical applications, such as
fraud detection, e.g., \cite{Branco20,BXu21}, and
healthcare, e.g., \cite{Morrison19,Park19}.
However, ML systems are threatened by adversarial attacks
\cite{Review2}.
Typically, ML systems involve a training stage and an inference stage \cite{Goodfellow_book}.
Accordingly, adversarial attacks can also be divided into \textit{training phase} attacks and \textit{inference phase} attacks \cite{sok_papernot}.

Inference phase attacks aim to either cause ML systems to produce adversary-selected outputs or collect evidence about the model characteristics \cite{sok_papernot}. For example, an \textit{adversarial evasion attack} uses carefully crafted adversarial examples (i.e., inputs with adversarial perturbations) to mislead ML models to make incorrect predictions \cite{sok_li,review,Szegedy_seminal,FGSM,DeepFool,CW}.
As another example, \textit{model extraction} attacks replicate the functional mapping rule of a victim ML model, an important asset of the model owner, by (e.g.) querying the victim model \cite{Reiter, Papernot3, orekondy19knockoff, Chandrasekaran2020}.

On the other hand, training phase attacks aim to subvert the model by modifying its training set. For example, data poisoning attacks degrade the prediction accuracy of a victim model by poisoning its training set with, e.g., mislabeled samples \cite{DP_SVM1, DP_SVM2, DP_DNN1, DP_DNN2, WildPatterns}. The backdoor attack focused on in this paper is also a type of training phase attack since the victim model will be planted with a backdoor.


\vspace{-0.1in}
\subsection{Classical Backdoor Attacks}\label{subsec:backdoor_attack}
\vspace{-0.1in}

A backdoor (Trojan) attack is an important type of training phase adversarial attack mainly targeting deep neural network classifiers. For a classification domain with sample space ${\mathcal X}$ and label space ${\mathcal Y}$, a classical backdoor attack aims to have the victim classifier $f:{\mathcal X}\rightarrow{\mathcal Y}$: (a) learn to classify to the attacker's target class $t\in{\mathcal Y}$, when a test sample ${\bf x}\in{\mathcal X}$ from any of the source classes ${\mathcal S}\subset{\mathcal Y}$ is embedded with the backdoor pattern; and (b) correctly classify clean test samples (without the backdoor pattern) \cite{BadNet}. 
For images (which is the most common domain of focus in the backdoor literature), common backdoor pattern types include: 1) an additive perturbation ${\bf v}$ embedded by 
$\tilde{\bf x} = [{\bf x} + {\bf v}]_{\rm c}$, 
where $||{\bf v}||_2$ is small (for imperceptibility) and $[\cdot]_{\rm c}$ is a domain-dependent clipping function, e.g., \cite{Targeted, Haoti, Post-TNNLS};
2) a local patch ${\bf u}$ embedded using an image-wide binary mask ${\bf m}$ by $\tilde{\bf x} = (1 - {\bf m})\odot{\bf x} + {\bf m}\odot{\bf u}$, where $||{\bf m}||_1$ is small for imperceptibility and $\odot$ represents element-wise multiplication, e.g.,
\cite{BadNet,NC};
and 3) a local or global pattern ${\bf u}$ ``blended'' using an image-wide binary mask ${\bf m}$ and a blending factor $\alpha\in(0, 1)$ by $\tilde{\bf x} = (1 - \alpha\cdot{\bf m})\odot{\bf x} + \alpha\cdot{\bf m}\odot{\bf u}$, where $\alpha$ is close to 0 for imperceptibility \cite{Targeted}.

The classical way to launch a backdoor attack, following the protocol of BadNet \cite{BadNet}, is by poisoning the training set of the classifier\footnote{Alternatively, backdoors may be planted by first intercepting the trained model and then directly modifying the model parameters \cite{Bai_2022_ECCV, bai2021targeted, Qi_2022_CVPR}.}.
The attacker first collects a small number of source class samples, embedded with the backdoor pattern, and then (re)labels them to the target class. Then, these created samples are inserted into the training set of the victim classifier for poisoning \cite{BadNet}. This classical backdoor attack is considered by most works on backdoor defense, including this paper. But unlike most prior works, our defense is also effective against more recent, advanced backdoor attacks (introduced shortly) launched by more capable attackers.

\vspace{-0.1in}
\subsection{Backdoor Defenses}\label{subsec:backdoor_defenses}
\vspace{-0.05in}

Backdoor defenses can be deployed during the classifier's training phase, post-training, or during inference. Each of these scenarios assumes a different defender role and capabilities.

Backdoor defenses during training aim to produce a backdoor-free classifier from the possibly poisoned training set \cite{SS, AC, Differential_Privacy, huang2022backdoor, trim_loss, shan2022poison, Hong20GS, doesntkill}. Existing defenses detect and remove suspicious samples from the training set \cite{SS, AC, shan2022poison}, identify outliers, and train only on non-outlier, trustworthy samples \cite{Differential_Privacy, huang2022backdoor, trim_loss},
or modify the training loss or training procedure for better robustness against data poisoning \cite{Hong20GS, doesntkill}.

Backdoor defenses deployed during the inference stage aim to detect test samples embedded with the backdoor pattern \cite{STRIP, voting, Februus, SentiNet, beatrix, scaleup}
and may also seek to correct the decisions on these samples. One type of method perturbs an input sample by overlapping a large number of benign samples or random noises and then uses the ensemble prediction results for detection \cite{STRIP, voting}. In \cite{Februus, SentiNet}, suspicious regions of input space are identified. 
In \cite{beatrix,InFlight}, the latent representation of clean samples is modeled, such that test-time inputs with the backdoor pattern will be detected as outliers.

In this paper, we focus on the {\it post-training} scenario, where the defender has the vantage point of a user of the classifier, with no access to the training set. A primary post-training defense task is backdoor detection, where the defender aims to detect if a given classifier is backdoor attacked \cite{TrojAI, Competition}. Existing defenses either i) reverse-engineer putative backdoor patterns for all classes and 
detect if some of these reverse-engineered patterns are anomalous with respect to
the others ({\it e.g.}, with an unusually small pattern size) 
\cite{NC,Tabor,Post-TNNLS,ABS,DataLimited, TwoClass, Shen2021BackdoorSF}; or ii) train a ``meta-classifier'' to recognize features extracted from the classifier being inspected \cite{meta_sup, META}. Another post-training defense task is backdoor mitigation, which aims to remove the learned backdoor mapping from a victim classifier \cite{TRC}. Mainstream approaches either fine-tune the classifier to ``unlearn'' the backdoor mapping \cite{NC, zeng2022adversarial, NAD} or prune neurons possibly activated by the backdoor pattern \cite{FP}.

\vspace{-0.1in}
\subsection{Advanced Backdoor Attacks}\label{subsec:backdoor_attack_advanced}
\vspace{-0.1in}

Recently, advanced backdoor attacks have been proposed to achieve better stealthiness against human inspectors. Clean-label attacks poison the training set of the classifier using samples collected from only the target class \cite{Clean_Label_BA, Hidden-trigger}.
These samples are perturbed to remove the original class-discriminating features and/or to embed features associated with the backdoor pattern.
There is no relabeling for these samples, which helps to avoid human suspicion during training.
In \cite{Haoti, Zhao_2022_CVPR, Wang_2022_CVPR}, the backdoor pattern is optimized (e.g. using a surrogate classifier) to be imperceptible to humans at test time.
Other strategies for achieving such test-time imperceptibility include embedding the backdoor pattern in the frequency domain \cite{wang2021backdoor} or leveraging a warping tragrantormation on the image to be embedded \cite{nguyen2021wanet}.
Alternatively, visible backdoor patterns can be embedded subtly, e.g., as a scene-plausible  physical object \cite{MAMF} or mimicking physical reflection on smooth surfaces \cite{Liu2020Refool}.

There are also advanced adaptive attacks proposed to evade particular types of backdoor defenses. For example, the ``all-to-all'' attack in \cite{BadNet}, can bypass backdoor defenses that assume a single target class, e.g. \cite{NC, Tabor}. The label-smoothed attack \cite{label_smoothed} is designed to bypass defenses based on backdoor pattern reverse-engineering \cite{NC, Post-TNNLS}. The Wasserstein-based backdoor attack in \cite{WB} is designed for clustering-based defenses deployed during the training phase, such as \cite{SS, AC, CI}.
The sample-specific backdoor attack proposed in \cite{li_ISSBA_2021} (a.k.a. input-aware backdoor \cite{nguyen2020inputaware}) can bypass defenses that assume a common backdoor pattern (in the input space), such as \cite{STRIP}.
However, most of these advanced backdoor attacks require strong adversary capabilities, as will be discussed next in Sec. \ref{sec:threat_model}. Even so, in our experiments, we will evaluate the detection capability of the proposed MM-BD against some representative advanced attacks mentioned above.

\vspace{-0.1in}
\section{Threat Model}\label{sec:threat_model}
\vspace{-0.05in}
\subsection{Attacker's Goals and Capabilities}\label{subsec:attacker_assumption}
\vspace{-0.05in}
In this paper, we focus on backdoor attacks against image classifiers. The extension of our method for other domains will be discussed in Sec. \ref{subsec:other_domain}. In particular, we consider three types of attackers with different levels of goals and capabilities to evaluate our proposed defense thoroughly.\\
{\bf Basic attacker:} An attack launched by a ``basic" attacker is a classical backdoor attack discussed in Sec. \ref{subsec:backdoor_attack}. The attack is associated with one backdoor target class, an arbitrary number of source classes, and a common backdoor pattern, such that samples from the source class are supposed to be misclassified to the target class when the backdoor pattern is present, with backdoor-free samples correctly classified. The attacker has the ability to poison the training set \cite{data_collection}. But the attacker has neither access to the samples originally in the classifier's training set, nor to the training process itself.
In this paper, we compare our proposed MM-BD and our mitigation approach with other methods that address basic attackers.\\
{\bf Advanced attacker:} These attackers are motivated by goals in addition to those for a basic attacker, such as the stealthiness of the attack against human inspection, and the evasiveness of the attack against backdoor defenses. Most of these additional goals cannot be achieved by a basic attacker. Thus, an advanced attacker is endowed with additional capabilities such as: (a) the capability to collect sufficient data and to train a surrogate classifier, and/or even (b) full control of the victim's training process. The latter is a particularly significant assumption which is valid only for a few scenarios, e.g., where there is an insider or training is outsourced to a third party which happens to be an attacker. In our experiments, MM-BD will be evaluated against several advanced backdoor attacks mentioned in Sec. \ref{subsec:advanced_attacks}.\\
{\bf Adaptive attacker:} The strongest attacker considered in this paper is an adaptive attacker who, in addition to the goals of a basic attacker, also aims to defeat mounted defenses. An adaptive attacker is stronger than the previous two types of attackers, with full control of the training process and full knowledge of the defense. In Sec. \ref{subsubsec:adaptive}, we create a strong, optimized adaptive attack based on these capabilities. We show that to bypass MM-BD, the attacker needs to solve a complex min-max optimization problem, which is time-consuming. 


\vspace{-0.1in}
\subsection{Defender's Goals and Assumptions}\label{subsec:defender_assumption}
\vspace{-0.1in}
In this paper, we consider the practical post-training defense scenario where the defense
is applied after the classifier has been trained, and without access to the training set \cite{ModelZoo}.
The most important goal for post-training defense is to detect if the classifier was backdoor-attacked. If an attack is detected, the user could choose a replacement classifier, {\it if} one is available. Otherwise, the backdoor attack should be mitigated, so that: a) the classifier predicts to the original source class when a test sample is embedded with the backdoor pattern, and b) the classification accuracy on clean, backdoor-free test samples is not significantly degraded.

The assumptions associated with the post-training defense scenario are summarized as follows:\\
(1) {\it The defender does not know {\it a priori} if there is an attack.} This is the fundamental assumption for the post-training backdoor detection problem to be meaningful.\\
(2) {\it No information about the backdoor pattern is available to the defender.} This assumption is relaxed by many existing post-training detectors -- THEY make assumptions about the backdoor pattern type, or how human-imperceptibility is achieved (e.g., a small $\ell_2$ norm for additive perturbation backdoor patterns in Sec. \ref{subsec:backdoor_attack}). However, MM-BD strictly {\it makes no assumptions} about the backdoor pattern.\\
(3) {\it The defender has no access to the classifier's training set.} The defender is a user of the classifier, or the user of a legacy system. In the former (proprietary) case, the training set of the classifier may not be publicly available, while in the latter (legacy) case it has long been forgotten.\\
(4) {\it There are no clean classifiers trained for the same domain.} Otherwise, the defender may directly use the clean classifier (possibly after some fine-tuning) in replacement of the classifier to be inspected. More importantly, post-training detection will be an easier ``semi-supervised'' problem if clean classifiers are available, since these classifiers can be used (e.g.) to set a conformal threshold for detection inference \cite{conformal}.\\
(5) {\it The defender is able to independently collect a small, clean data set containing samples from all classes in the domain.} This assumption is used by most post-training detectors, e.g., \cite{NC,Tabor,Post-TNNLS,ABS, TwoClass, Shen2021BackdoorSF}.
However, MM-BD {\it does not need} any clean samples for detection. This makes MM-BD applicable to scenarios where clean samples are very rare or expensive.

\vspace{-0.05in}
\section{Related Work}\label{sec:related_work}
\vspace{-0.05in}
{\bf Post-training backdoor detection methods.} Existing methods include a family of ``meta-learning'' approaches, where a large number of shadow neural networks, labeled ``attack'' or ``no attack'', are trained -- features extracted from these labeled networks are treated as supervised examples, input to a binary ``meta-classifier'' trained to discriminate ``attack'' from ``no attack'' \cite{meta_sup, META}. However, these methods employ a pool of backdoor pattern types, and may fail when the actual type used by the attacker is not in the pool. Moreover, training the shadow networks requires a relatively large number of clean samples and is heavily computational. Another family of detectors trial reverse-engineer the backdoor pattern for each putative target class \cite{NC, Tabor, Post-TNNLS, ABS, DataLimited, TwoClass, UMD}. Such reverse-engineering is performed using a small set of clean samples \cite{NC}, or using simulated samples obtained by model inversion \cite{DeepInspect, model_inversion}. Detection inference is then based on statistics obtained from the estimated backdoor patterns (e.g. the $\ell_1$ norm of the estimated mask for patch replacement backdoor patterns). However, reverse-engineering relies on knowledge of the backdoor pattern type\cite{NC, Post-TNNLS, META}. Also, most of these reverse-engineering-based methods, except \cite{DeepInspect, model_inversion}, require some clean samples, which are not always available as we have discussed in Sec. \ref{subsec:defender_assumption}. In comparison, MM-BD does not rely on knowledge of the backdoor pattern type and does not need any clean samples. Moreover, some reverse-engineering-based methods fail when the backdoor attack involves only a few source classes, as shown in\cite{Post-TNNLS, Shen2021BackdoorSF}. \cite{Post-TNNLS} addressed this issue with a significant increment in computational complexity. While \cite{Shen2021BackdoorSF}
estimates the source classes and the target class via a complicated optimization procedure, our method can accurately detect backdoor attacks with an arbitrary number of source classes and is computationally efficient, as will be shown in our experiments. 


{\bf Post-training backdoor mitigation methods.} In \cite{NC, zeng2022adversarial, NAD}, a detected classifier is fine-tuned on a large number of clean samples to ``unlearn'' the backdoor mapping. In \cite{FP, zheng2022data}, neurons possibly associated with the backdoor pattern are pruned based on their activation or modulus of (Lipschitz) continuity. Although the main focus of our paper is backdoor detection, in Sec. \ref{subsec:backdoor_mitigation}, we also propose a method that mitigates backdoors, for most types of backdoor patterns, using very few clean samples, and without fine-tuning any of the classifier's original parameters. Our method can also be combined with existing ones to achieve even better performance.

{\bf Backdoor defenses addressing various backdoor pattern types.} There are many backdoor defenses capable of addressing various backdoor pattern types. In particular, most training phase backdoor defenses, such as \cite{SS, AC, Differential_Privacy, huang2022backdoor, trim_loss, shan2022poison, Hong20GS, doesntkill}, and some inference-stage backdoor defenses, such as
\cite{beatrix,STRIP,InFlight}, are not designed with any particular type of backdoor pattern in mind (i.e., they are backdoor agnostic). 
However, for the post-training backdoor detection problem focused on in this paper, the defender has no access to any actual backdoor patterns, which is different from the previous two defense scenarios. \tunedown{Thus, some post-training detectors tend to rely on assumptions about the backdoor pattern incorporation mechanism.} For example, \cite{NC} addresses patch-based patterns, \cite{Post-TNNLS} assumes imperceptible perturbation patterns, and \cite{Clear} considers attacks that do not even involve a backdoor pattern. In contrast to these, our post-training detector makes no assumptions about the incorporation mechanism of the backdoor pattern.

\vspace{-0.1in}
\section{Method}\label{sec:method}
\vspace{-0.1in}

In this section, we first discuss the key ideas behind our detector in Sec. \ref{subsec:key_ideas}. In Sec. \ref{subsec:detection_procedure} we present our detection procedure, which consists of an estimation step to obtain a novel {\it maximum margin} (MM) statistic for each class, followed by unsupervised detection inference. We then propose a mitigation method in Sec. \ref{subsec:backdoor_mitigation}.

\vspace{-0.1in}
\subsection{Key Ideas}\label{subsec:key_ideas}
\vspace{-0.1in}

Unlike existing detectors based on the small patch size (\cite{NC}) or the small perturbation size (\cite{Post-TNNLS}) of the backdoor pattern, our detector is based on the influence of a backdoor attack on the landscape of the classifier's logits $\{g_c(\cdot) | c\in  {\mathcal Y}\}$ (i.e., the activations directly before the softmax layer), independent of the backdoor pattern type. Consider a backdoor attack with target class $t\in{\mathcal Y}$ and denote the classifier's logit function associated with any class $c\in{\mathcal Y}$ as $g_c:{\mathcal X}\rightarrow{\mathbb R}$ (assuming ${\mathcal X}$ compact without loss of generality). For all $c\in{\mathcal Y}\setminus t$, regardless of the backdoor pattern type, we will likely observe:
\begin{equation}\label{eq:key_idea}
\begin{split}
\max_{{\bf x}\in{\mathcal X}} \big[ g_t({\bf x}) - &\max_{k\in{\mathcal Y}\setminus t} g_{k}({\bf x})\big] \gg  \\&\max_{{\bf x}\in{\mathcal X}} \big[ g_c({\bf x}) - \max_{k'\in{\mathcal Y}\setminus c} g_{k'}({\bf x})\big]
\end{split}
\end{equation}
I.e., the {\it maximum margin statistic} for the true backdoor attack target class (defined as the left hand side of \eqref{eq:key_idea}) will tend to be {\it much larger} than the maximum margin statistics for all other classes (right hand side of \eqref{eq:key_idea}).

\begin{figure*}[t]
\vspace{-0.4in}
	\centering
	\begin{minipage}[b]{.26\linewidth}
		\centering
		\centerline{\includegraphics[width=1.15\linewidth]{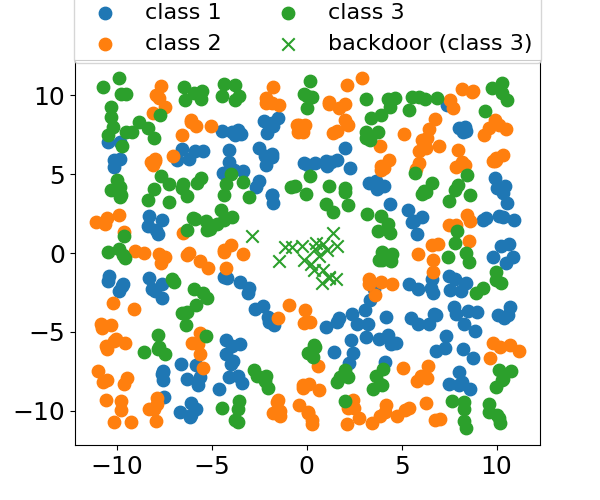}}
		\subcaption{Illustration of the distribution of samples from all three classes.}
	\end{minipage}
	\begin{minipage}[b]{.23\linewidth}
		\centering
		\centerline{\includegraphics[width=1.05\linewidth]{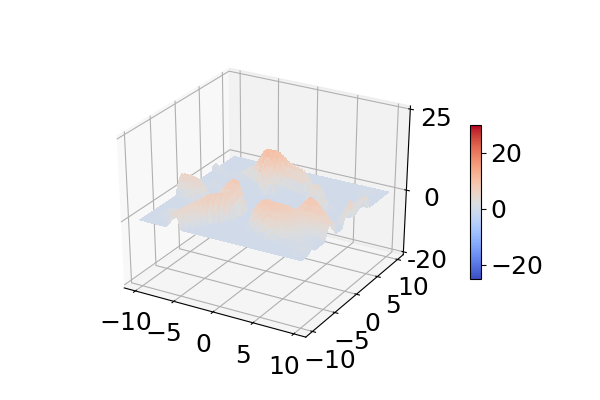}}
		\subcaption{BA10-class1}
		\centerline{\includegraphics[width=1.05\linewidth]{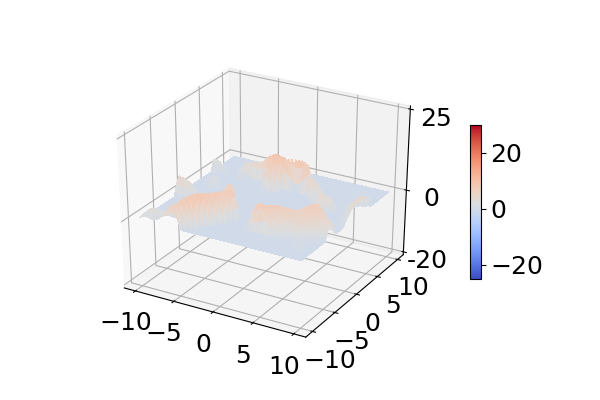}}
		\subcaption{BA100-class1}
	\end{minipage}
	\begin{minipage}[b]{.23\linewidth}
		\centering
		\centerline{\includegraphics[width=1.05\linewidth]{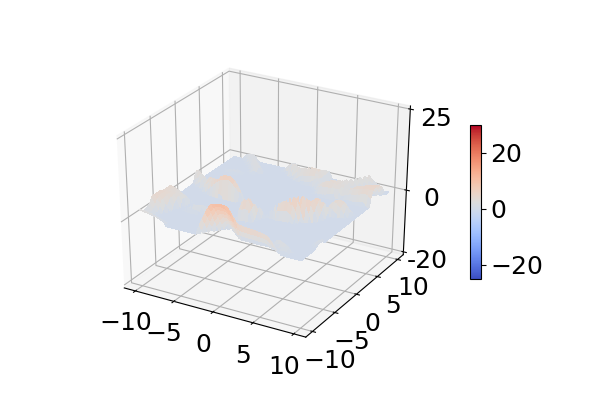}}
		\subcaption{BA10-class2}
		\centerline{\includegraphics[width=1.05\linewidth]{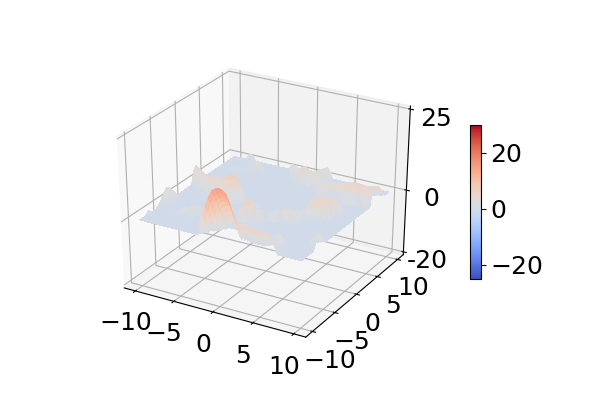}}
		\subcaption{BA100-class2}
	\end{minipage}
	\begin{minipage}[b]{.23\linewidth}
		\centering
		\centerline{\includegraphics[width=1.05\linewidth]{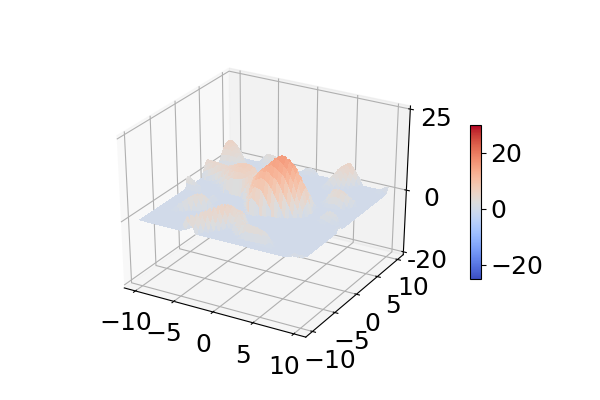}}
		\subcaption{BA10-class3}\label{subfig:BA10_class3}
		\centerline{\includegraphics[width=1.05\linewidth]{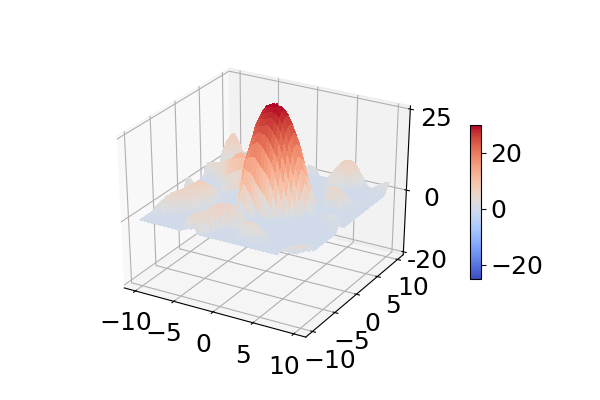}}
		\subcaption{BA100-class3}\label{subfig:BA100_class3}
	\end{minipage}
	\caption{For two-dimensional input and three classes following the sample distribution in (a), consider two backdoor attacks (BA-10 and BA-100) with different poisoning rates but the same target class 3. For both backdoor attacks, the maximum margin for target class 3 ((f) and (g)) is abnormally large compared with that for the non-target classes ((b), (c), (d), (e)).\vspace{-0.2in}}
	\label{fig:simulation}
\end{figure*}

{\it Why a backdoor attack causes the above phenomenon?} Note that a backdoor pattern is a common pattern\footnote{Some recently proposed advanced backdoor attacks use backdoor patterns that are sample-specific in the input space \cite{nguyen2020inputaware, li_ISSBA_2021} (Sec. \ref{subsec:backdoor_attack_advanced}). However, these backdoor patterns embedded in different samples still share common semantic features and are similar to each other in some latent embedding space. Thus, these backdoor attacks are still detectable by our method, as will be shown in Sec. \ref{subsec:advanced_attacks}.} embedded in all samples used for poisoning the classifier's training set; and the same backdoor pattern will be embedded in test samples to induce test-time misclassifications. By contrast, class-discriminating features not associated with the backdoor pattern typically exhibit high variability\footnote{In extreme cases, class-discriminating features can be highly common across samples from the same class, which creates an ``intrinsic" backdoor that is hardly distinguishable from backdoors planted by an attacker \cite{PCBD}. A general solution to rule out such intrinsic backdoors is still an open problem.} -- e.g., 
a `bird' category with multiple bird species, or even the same object captured under different views, range, or lighting conditions. The commonality of a backdoor pattern is critical for it to override class-discriminating features that will favor deciding to the source class and so that the backdoor pattern can be easily learned (at a low poisoning rate \cite{BAsurvey}) by the victim classifier during training. However, the repetition\footnote{We acknowledge a severely imbalanced training set may cause a similar overfitting phenomenon, which may lead to false detections. However, an imbalanced training set itself is harmful to the classifier, as will be discussed in detail in Sec. \ref{subsubsec:class_imbalance}.} of the common backdoor pattern in the training set also induces an inevitable overfitting which: a) boosts the target class logit (by causing abnormally large activations from neurons positively correlated with this logit), and b) suppresses the logits of all other classes (as will be shown empirically in Sec. \ref{subsubsec:ablation}). Consequently, an abnormally large margin between the target class logit and logits of all other classes will be created, due to both the ``boosting'' and the ``suppression'' effects.

It is worth mentioning that the concurrent work \cite{sensitivity} also reveals that the internal feature representation of backdoor samples will be dominated by the backdoor pattern rather than by the benign class-discriminative features, and such domination is mainly caused by overfitting. However, \cite{sensitivity} focuses on during-training defense, with the training set available, while our work is post-training, with the training set unavailable.


The above reasoning is illustrated in Fig. \ref{fig:simulation} using a toy, visualizable example. We consider two-dimensional inputs from three classes, with the sample distribution for each class specified by a Gaussian mixture model (with a large number of components to mimic the high variability of class-discriminating features). We generated 500 training samples per class, and launched two backdoor attacks (both with target class 3), respectively with 10 (BA-10) and 100 (BA-100) ``backdoor samples'' inserted for poisoning. For each backdoor attack, we trained a multilayer perceptron with three hidden layers, which achieves nearly 91\% accuracy on clean test samples for both backdoor attacks. The attack success rates of the trained classifiers on test ``backdoor samples'' are 92\% and 99\% for BA-10 and BA-100, respectively. In Fig. \ref{fig:simulation}, for each backdoor attack and each class, we plot the margin between a class's logit and the largest logit among the other two classes as a function of the input space (only positive values are kept for better visualization). We observe that the backdoor target class (Fig. \ref{subfig:BA10_class3} and \ref{subfig:BA100_class3}) has abnormally large maximum margin for both backdoor attacks (which agrees with Eq. \eqref{eq:key_idea}). Moreover, compared with BA-10, the atypicality of the maximum margin for the target class is more obvious for BA-100 (note that the higher poisoning rate may be preferred by the attacker since this yields a higher attack success rate).

To give another perspective on our hypothesis,
consider a simple linear model $g(\cdot): {\mathcal X}\rightarrow{\mathbb R}^{|{\mathcal Y}|}$ where
the output related to class $c \in {\mathcal Y}$ is given by the logit $g_c$
formed by the column-vector dot-product,
$g_c(x) = {\bf w}_c' {\bf x}$, 
where: ${\bf w}_c$ represents the 
weight vector related to class $c$, which has the same dimension as input ${\bf x}$; 
and
the class  decision $c$ of the model for ${\bf x}$ has largest associated logit $g_c({\bf x})$.
Also assume that, after training, the 
model can correctly classify all the training samples with 
confidence higher than $\tau$, assessed by the margin of the correct class 
(which is the 
logit for that class minus the maximum logit among all other classes):
\begin{equation}
\label{eq:margin}
\begin{split}
    g_{y}({\bf x})  - \max_{c \neq y} g_c({\bf x}) & = {\bf w}_{y}' {\bf x}  - \max_{c \neq y} {\bf w}_c ' {\bf x} \\
    &\geq \tau, ~~\forall ({\bf x},y) \in {\mathcal X}\times {\mathcal Y}.
\end{split}
\end{equation}
Suppose a backdoor pattern ${\bf v}$ is additively incorporated 
into a clean training image ${\bf x}_s$, which is originally from source-class $s \neq t$ and relabeled to target class $t$. Eq.  \eqref{eq:margin} implies:
\begin{equation}
\label{eq:attack_margin}
     g_t({\bf x}_s + {\bf v}) - g_s({\bf x}_s + {\bf v}) = {\bf w}_t '({\bf x}_s + {\bf v}) -  {\bf w}_s '({\bf x}_s + {\bf v}) \geq \tau.
\end{equation}
Based on the confident margin assumption in Eq. \eqref{eq:margin}, for the clean $x_s$ we have
\begin{equation}
\label{eq:clean_margin}
     {\bf w}_s'{\bf x}_s - {\bf w}_t'{\bf x}_s \geq \tau, 
\end{equation}
So, adding Eq. (\ref{eq:attack_margin}) and Eq. (\ref{eq:clean_margin}) gives:
\begin{equation}
     g_t({\bf v}) -g_s({\bf v}) = 
     {\bf w}_t '{\bf v} - {\bf w}_s' {\bf v} \geq 2\tau.
\end{equation}
That is, the lower bound of the maximum margin for the backdoor target class is at least $2\tau$, {\it larger} than the lower bound for legitimate samples from all classes.

\vspace{-0.1in}
\subsection{Detection Procedure}\label{subsec:detection_procedure}
\vspace{-0.1in}

{\bf Estimation step.} For each class $c\in{\mathcal Y}$, we estimate a maximum margin statistic by solving:
\begin{equation}\label{eq:opt_main}
\maximize_{{\bf x}\in{\mathcal X}} \quad g_c({\bf x}) - \max_{k\in{\mathcal Y}\setminus c} g_{k}({\bf x})
\end{equation}
using gradient ascent (common for model inversion \cite{model_inversion} and related applications, e.g., generating adversarial examples \cite{PGD}) with projection onto ${\mathcal X}$ (e.g. ${\mathcal X}=[0, 1]^{H\times W\times C}$ for color images with height $H$, width $W$, and $C$ channels).

Note that ${\mathcal X}$ is a closed convex set; thus, the continuous logit function on ${\mathcal X}$ is bounded and Lipschitz.
In other words, there exist closed balls in which the logit function is convex and with a local maximum.
Then, according to Theorem 3.2 in \cite{bubeck2014convex}, convergence is guaranteed for gradient ascent from any random initialization in ${\mathcal X}$ with projection onto ${\mathcal X}$.
As is common practice, we perform multiple random initializations in ${\mathcal X}$ (e.g., for images, pixel values are uniformly randomly initialized in the interval $[0, 1]$), and pick the largest local optimal solution. Compared with reverse-engineering-based defenses that may suffer from mismatch between the assumed backdoor pattern embedding type and the true attack backdoor pattern type, our optimization problem does not require assuming a backdoor pattern embedding type. Moreover, reverse-engineering-based defenses' backdoor pattern estimation using clean samples from {\it all} non-target classes has been found experimentally to fail when the majority of these classes are not source classes 
\cite{Shen2021BackdoorSF}. By contrast, our method can detect backdoor attacks with an arbitrary number of source classes and does not require knowledge of any legitimate samples from the domain.

{\bf Detection inference step.} We propose an unsupervised anomaly detector as given in \cite{Post-TNNLS}. Denote the estimated maximum margin statistic for each class $c\in{\mathcal Y}$ as $r_c$ and the largest statistic as $r_{\rm max}=\max_{c\in{\mathcal Y}} r_c$. We hypothesize that, when there is a backdoor attack, $r_{\rm max}$ will be associated with the target class and will be an outlier to the distribution of maximum margin statistics for non-target classes. Thus, we estimate a null distribution $H_0$ using all statistics excluding $r_{\rm max}$. Given that the maximum margin is strictly positive (both theoretically and empirically for the estimated maximum margin in our experiments), we choose single-tailed density forms for the null distribution, e.g. Gamma distribution, in our experiments. In order to evaluate the atypicality of $r_{\rm max}$ under the estimated null, we compute an order-statistic p-value:
\begin{equation}
\label{eq:pv}
{\rm pv}=1-H_0(r_{\rm max})^{K-1},
\end{equation}
where $K=|{\mathcal Y}|$ is the total number of classes in the domain. It can easily be shown that ${\rm pv}$ follows a uniform distribution on $[0, 1]$ under the null hypothesis of ``no attack''. Thus, we claim a detection with confidence $1-\theta$ (e.g. with the classical $\theta=0.05$) if ${\rm pv}<\theta$. If a backdoor attack is detected, the class associated with $r_{\rm max}$ is inferred as the backdoor target class.

\vspace{-0.1in}
\subsection{Mitigation of Backdoor Attacks}\label{subsec:backdoor_mitigation}
\vspace{-0.1in}

When a backdoor attack is detected, 
predictions made by the victim classifier to classes other than the detected target class, or predictions made on the test examples that are not embedded with backdoor trigger might still be trustworthy. An option is to {\it mitigate} the backdoor attack.

Our mitigation approach is based on the observation that a backdoor attack induces a subset of neurons in each layer to have abnormally large activations. Such ``large activation'' phenomenon is also the basis of the backdoor {\it detection} approach in \cite{ABS}, though several hyperparameters are required, e.g., to identify the neurons responsible for the large activations. In contrast, we apply to {\it every neuron} a specific optimized upper bound in order to suppress any possible large activations caused by a backdoor attack, without significant degradation in the classifier's accuracy on clean samples. Let $\sigma_l:{\mathbb{R}^{n_{l-1}}}\rightarrow{\mathbb{R}^{n_l}}$ be the activations of the $l$-th layer (for $l=1, \cdots, L$) of the victim classifier (as a function of the activations of the previous layer). Here, we do not explicitly represent the parameters in each layer for brevity, since none of them will be modified during our {\it mitigation} process. Then, the logit function for any class $c\in{\mathcal Y}$ and any input ${\bf x}\in{\mathcal X}(={\mathbb R}^{n_0})$ can be written as:
\begin{equation}\label{eq:logit}
g_c({\bf x}) = {\bf w}^T_c (\sigma_L \circ \cdots \circ \sigma_1 ({\bf x}))+b_c,
\end{equation}
where ${\bf w}_c\in{\mathbb R}^{n_L}$ and $b_c\in{\mathbb R}$ are the weight vector and bias associated with class $c$ respectively. For each layer $l=2, \cdots, L$, we also denote a bounding vector ${\bf z}_l\in{\mathbb R}^{n_l}$, such that the logit function, {\it with bounded activation}, for each class $c\in{\mathcal Y}$ and any input ${\bf x}$ can be represented by:
\begin{equation}\label{eq:logit_bounded}
\begin{split}
\bar{g}_c({\bf x}; {\bf Z}) &  = {\bf w}^T_c ( \bar{\sigma}_L ( \bar{\sigma}_{L-1} ( \cdots \bar{\sigma}_2( \sigma_1 ({\bf x}) \\ & ;{\bf z}_2) \cdots ;{\bf z}_{L-1}) ;{\bf z}_L) )+b_c,
\end{split}
\end{equation}
where ${\bf Z}=\{{\bf z}_2, \cdots, {\bf z}_L\}$ and $\bar{\sigma}_l(\cdot;{\bf z}_l)=\min\{\sigma_l(\cdot), {\bf z}_l\}$ for any $l=2, \cdots, L$ (and where the `min' operator is applied to each component of the vector). To find the minimum activation upper bound for each neuron without affecting the classifier's performance on clean test samples, we propose to solve the following problem on a small set ${\mathcal D}$ of clean samples:
\begin{equation}\label{eq:opt_mitigation}
\begin{aligned}
	& \underset{{\bf Z}=\{{\bf z}_2, \cdots, {\bf z}_L\}}{\text{min}}
	\ \  \sum_{l=2}^L ||{\bf z}_l||_2 \\
	& \text{subject to} \ \ 
	 \frac{1}{|{\mathcal D}|}\sum_{({\bf x}, y)\in{\mathcal D}}{\mathds 1}[y=\argmax_{c\in{\mathcal Y}} \bar{g}_c({\bf x}; {\bf Z})] \geq \pi,
\end{aligned}
\end{equation}
where ${\mathds 1}[\cdot]$ represents the indicator function, and $\pi$ is the minimum accuracy benchmark (e.g., set $\pi=0.95$). Here, we minimize the $\ell_2$ norm of the bounding vectors to penalize activations with overly large absolute values in each layer. To practically solve the above problem, we propose to minimize the following Lagrangian using gradient descent:
\begin{equation}\label{eq:lagrangian_mitigation}
\begin{split}
L({\bf Z}, \lambda; {\mathcal D}) &= \frac{1}{|{\mathcal D}|\times|{\mathcal Y}|} \sum_{({\bf x}, y)\in{\mathcal D}} \sum_{c\in{\mathcal Y}} [\bar{g}_c({\bf x}; {\bf Z})  \\ & - g_c({\bf x})]^2 
+ \lambda \sum_{l=2}^L ||{\bf z}_l||_2,
\end{split}
\end{equation}
where ${\bf Z}$ is initialized large. The first term of Eq. \eqref{eq:lagrangian_mitigation} aims to keep the classifier's logits for the samples in ${\mathcal D}$ unchanged. Such a design not only helps satisfy the accuracy constraint in problem \eqref{eq:opt_mitigation}, but also avoids having the logit associated with the class label for a given sample to further grow (i.e. overfit), allowing mitigation to be achieved with limited samples\footnote{Other choices, such as a differentiable surrogate for correct classification count, might lead to overfitting.}. $\lambda$ is updated automatically to fulfill the constraint of problem \eqref{eq:opt_mitigation}. This process is summarized in Alg. \ref{alg:mitigation}. 
Note that 
we initialize the values in ${\bf Z}$ large enough such that no activation bounding/saturation is initially performed.
This can be easily achieved by feeding in clean samples to get a rough range for the activations, and then setting the initial upper bound to a magnitude larger than typical activations.

Finally, a class posterior with the backdoor mitigated is obtained by applying a softmax to the logits $\{\bar{g}_c({\bf x}; {\bf Z}^{\ast})\}_{c\in{\mathcal Y}}$.

\vspace{-0.1in}
\section{Experiments}\label{sec:experiments}
\vspace{-0.1in}
Experiments are conducted mainly on four benchmark datasets with different image resolutions, image sizes and numbers of classes: CIFAR-10, CIFAR-100 \cite{CIFAR10}, TinyImageNet, and GTSRB \cite{GTSRB} (with more details in Apdx. \ref{sec:detail_dataset}).


\vspace{-0.05in}
\subsection{Main Experiments for Backdoor Detection}\label{subsec:exp_main}
\vspace{-0.05in}

In the following, we show the effectiveness of our backdoor detector MM-BD in terms of {\it detection accuracy} and {\it computational efficiency} compared with several state-of-the-art post-training detectors for a variety of backdoor attack configurations (consisting of the backdoor pattern type and the number of source classes). While here we focus on classical backdoor attacks commonly considered by existing backdoor detection works and recent competitions like \cite{Competition}, the success of MM-BD against many other advanced backdoor attack settings and adaptive backdoor attacks is shown in Sec. \ref{subsec:advanced_attacks}.

\vspace{-0.1in}
\subsubsection{Settings}\label{subsubsec:settings}

We consider three common backdoor embedding types from the backdoor attack literature: {\it additive}, {\it patch replacement}, and {\it blending} (with associated embedding functions in Sec. \ref{subsec:backdoor_attack}). Effectiveness of MM-BD on warping-based backdoor patterns and sample-specific backdoor patterns is shown in Sec. \ref{subsec:advanced_attacks}. In particular, we consider the following five respective backdoor patterns. For the additive backdoor pattern, we consider a {\it global} ``chessboard'' pattern from \cite{Post-TNNLS} and a {\it local} ``1-pixel'' perturbation from \cite{SS}. For the patch replacement backdoor pattern type, we consider a BadNet pattern from \cite{BadNet} and a ``{\it unicolor}'' patch from \cite{NC}. For the blended backdoor pattern type, we consider a ``blended'' {\it noisy} patch from \cite{Post-TNNLS}. Examples of these backdoor patterns and more details are in Apdx. \ref{sec:detail_bp}.



Like most existing works (e.g. \cite{NC, Post-TNNLS, Tabor, ABS}), we consider backdoor attacks with one target class (randomly selected for each attack). However, we allow one or multiple source classes -- for brevity, these two source class settings are denoted as `S' (for ``single'') and `M' (for ``multiple'') respectively. For each backdoor attack with the `S' setting, the source class is randomly selected. For TinyImageNet, we randomly select ten source classes for each backdoor attack with the `M' setting; while for the other three datasets, all classes other than the target class are selected as source classes (which is assumed by the detectors in \cite{NC, Tabor, ABS}) for the `M' setting. With the above notations, a backdoor attack with ``chessboard'' backdoor pattern and multiple source classes is denoted `chessboard-M'.

For CIFAR-10, we created ten ensembles of backdoor attacks for all ten combinations of 5 different backdoor patterns and `S'/`M' respectively. For CIFAR-100 and GTSRB, we created five backdoor attack ensembles for 5 backdoor patterns respectively, all with the `M' setting. We did not create backdoor attacks with a single source class for these two datasets because we cannot generate sufficient backdoor training images from a single class to launch a successful backdoor attack, given the limited number of images in each class. For TinyImageNet, we only generated one backdoor attack ensemble with the setting `BadNet-M' (which is arbitrarily selected) since the amount of time just to train a classifier on this dataset is extraordinarily high. For each ensemble, ten different attacks were generated independently according to the specified settings using the classical ``data poisoning'' protocol in \cite{BadNet}. Other configurations, including the number of backdoor training images created for backdoor attacks in each ensemble, are deferred to Apdx. \ref{sec:ba_config}.

We trained one classifier for each backdoor attack. The DNN architectures for backdoor attacks on CIFAR-10, CIFAR-100, TinyImageNet, and GTSRB were ResNet-18 \cite{ResNet}, VGG-16 \cite{VGG}, ResNet-34 \cite{ResNet}, and MobileNet \cite{MobileNetV2}, respectively. 
For each dataset, we also created an ensemble of clean classifiers to evaluate the false detection rate. All the backdoor attacks we created are successful with high attack success rate (ASR) and negligible degradation in clean test accuracy (ACC)\footnote{ASR and ACC are commonly used for assessing the effectiveness of backdoor attacks; see definitions in \cite{Post-TNNLS, DataLimited}.}, compared with the clean classifiers trained for the same dataset. More details are shown in Apdx. \ref{sec:training_config}.


\vspace{-0.05in}
\subsubsection{Detection Performance}\label{subsubsec:detection_performance}

We compare MM-BD with six state-of-the-art post-training detection methods: NC \cite{NC}, TABOR \cite{Tabor}, ABS \cite{ABS}, PT-RED\cite{Post-TNNLS}, META \cite{META}, and TND\cite{DataLimited}. We followed the original implementations of these methods with only modest changes (e.g. choosing the best detection threshold to maximize their performance). Especially for META, we used the official code to train the ``meta-classifier'' for detection. For MM-BD, we solved problem \eqref{eq:opt_main} using gradient ascent with convergence criterion\footnote{Stopping when the relative change of the objective function between any two consecutive iterations is less than $\epsilon$.} $\epsilon=10^{-5}$ and 30 random initializations. These choices are not critical to the detection accuracy. For the inference stage, the detection threshold was set to $\theta=0.05$, i.e., 0.95 detection confidence, which is the classical threshold for statistical hypothesis tests and was kept fixed in all experiments in this paper.


\begin{table*}[!t]
\vspace{-0.2in}
	\scriptsize
	\begin{center}
 \setlength{\tabcolsep}{2.5pt}
			\begin{tabular}{ccccccccccccc}
				\toprule
				& & \multicolumn{11}{c}{CIFAR-10}        \\ \cmidrule(lr){3-13} 
				&$N_{\rm img}$ & clean & chessboard-S  & chessboard-M & 1-pixel-S& 1-pixel-M& BadNet-S &BadNet-M &unicolor-S&unicolor-M&blend-S&blend-M\\
				\midrule
				NC\cite{NC}&10 &56.7$\pm$49.6&66.7$\pm$47.1& \cellcolor{green!60} 100.0$\pm$0.0&56.7$\pm$49.6&36.7$\pm$48.2& \cellcolor{green!60} 93.3$\pm$24.9& \cellcolor{green!60} 96.7 $\pm$ 18.0& \cellcolor{green!60} 90.0$\pm$30.0&76.7$\pm$42.3&36.7$\pm$48.2&33.3$\pm$47.1\\
				TABOR\cite{Tabor}&10&70.0$\pm$45.8&70.0$\pm$45.8& \cellcolor{green!60} 93.3$\pm$24.9&73.3$\pm$44.2& \cellcolor{green!60} 93.3$\pm$24.9&63.3$\pm$48.2&73.3$\pm$44.2&26.7$\pm$44.2&66.7$\pm$47.1&73.3$\pm$44.2&\cellcolor{green!15} 86.7$\pm$34.0\\
				ABS\cite{ABS}&1 &\cellcolor{green!60} 93.3$\pm$24.9&26.7$\pm$44.2&66.7$\pm$47.1&66.7$\pm$47.1&\cellcolor{green!15} 80.0$\pm$40.0& 46.7$\pm$49.9& \cellcolor{green!60} 96.7$\pm$18.0&73.3$\pm$44.2&46.7$\pm$49.9&\cellcolor{green!15} 80.0$\pm$40.0&\cellcolor{green!60} 90.0$\pm$30.0\\
		META\cite{META}&10k& 73.3$\pm$44.2& 76.7$\pm$42.3&60.0$\pm$49.0&10.0$\pm$30.0&6.7$\pm$24.9& \cellcolor{green!15} 83.3$\pm$37.3& \cellcolor{green!15} 83.3$\pm$37.3&30.0$\pm$45.8&30.0$\pm$45.8& 76.7$\pm$42.3 &\cellcolor{green!15} 86.7$\pm$34.0\\ 
				TND\cite{DataLimited} &5&50.0$\pm$50.0&10.0$\pm$30.0&26.7$\pm$44.2&46.7$\pm$49.9& \cellcolor{green!15} 83.3$\pm$37.3&20.0$\pm$40.0&33.3$\pm$47.1&6.7$\pm$24.9&3.3$\pm$18.0&40.0$\pm$49.0&56.7$\pm$49.6\\
				PT-RED\cite{Post-TNNLS} &100&76.7$\pm$42.3&\cellcolor{green!60} 100.0$\pm$0.0&\cellcolor{green!60} 96.7$\pm$18.0&\cellcolor{green!60} 93.3$\pm$24.9& \cellcolor{green!60} 100.0$\pm$0.0&23.3$\pm$42.3&13.3$\pm$34.0&16.7$\pm$37.3&23.3$\pm$42.3&46.7$\pm$49.9&63.3$\pm$48.2\\
				PT-RED+ABS &100& 66.7$\pm$47.1&\cellcolor{green!60} 100.0$\pm$0.0 &  \cellcolor{green!60} 100.0$\pm$0.0 & \cellcolor{green!60} 93.3$\pm$24.9&  \cellcolor{green!60} 100.0$\pm$0.0 &56.7$\pm$49.6& \cellcolor{green!60} 96.7$\pm$18.0&76.7$\pm$42.3 &56.7$\pm$49.6 & \cellcolor{green!15} 86.7$\pm$34.0 &\cellcolor{green!60} 96.7$\pm$18.0\\
				MM-BD(ours)& 0& \cellcolor{green!15} 86.7$\pm$34.0& \cellcolor{green!15} 86.7$\pm$34.0& \cellcolor{green!60} 90.0$\pm$30.0& \cellcolor{green!15} 83.3$\pm$37.3& \cellcolor{green!60} 100.0$\pm$0.0& \cellcolor{green!60} 90.0$\pm$30.0& \cellcolor{green!60} 100.0$\pm$0.0& 76.7$\pm$42.3& \cellcolor{green!60} 100.0$\pm$0.0& \cellcolor{green!15}\cellcolor{green!15} 86.7$\pm$34.0& \cellcolor{green!60}\cellcolor{green!60} 100.0$\pm$0.0\\
				\hline 
				
				& & \multicolumn{4}{c}{CIFAR-100}& \multicolumn{2}{c}{TinyImageNet} &   \multicolumn{5}{c}{GTSRB}   \\ \cmidrule(lr){3-6} \cmidrule(lr){7-8}\cmidrule(lr){9-13}
				& & clean & 1-pixel-M  & BadNet-M & unicolor-M & clean & BadNet-M &clean &chessboard-M&1-pixel-M&unicolor-M&blend-M
				\\
				\midrule
				NC&1&53.3$\pm$49.9&40.0$\pm$49.0&\cellcolor{green!60} 93.3$\pm$24.9& \cellcolor{green!60} 90.0$\pm$30.0&30.0$\pm$45.8& \cellcolor{green!15} 80.0$\pm$40.0&66.7$\pm$47.1& \cellcolor{green!60} 90.0$\pm$30.0&66.7$\pm$47.1&20.0$\pm$40.0&63.3$\pm$48.2\\
				TABOR&1&43.3$\pm$49.6&70.0$\pm$45.8&60.0$\pm$49.0&56.7$\pm$49.6&40.0$\pm$49.0&70.0$\pm$45.8&56.7$\pm$49.6&\cellcolor{green!60} 90.0$\pm$30.0&\cellcolor{green!15}80.0$\pm$40.0&20.0$\pm$40.0&33.3$\pm$47.1\\
				ABS&1& \cellcolor{green!60} 96.7$\pm$18.0&20.0$\pm$40.0&\cellcolor{green!15} 86.7$\pm$34.0& \cellcolor{green!60} 90.0$\pm$30.0& \cellcolor{green!60} 90.0$\pm$30.0&20.0$\pm$40.0& \cellcolor{green!15} 83.3$\pm$37.3&30.0$\pm$45.8& 76.7$\pm$42.3&56.7$\pm$49.6&60.0$\pm$49.0\\
				TND&1&26.7$\pm$44.2&26.7$\pm$44.2&23.3$\pm$42.3&20.0$\pm$40.0&50.0$\pm$50.0&30.0$\pm$45.8&60.0$\pm$49.0&20.0$\pm$40.0&30.0$\pm$45.8&0.0$\pm$0.0&10.0$\pm$30.0\\
    PT-RED&2&\cellcolor{green!60}90.0$\pm$30.0&\cellcolor{green!60} 93.3$\pm$24.9&33.3$\pm$47.1&20.0$\pm$40.0&\cellcolor{green!60} 100.0$\pm$0.0&0.0$\pm$0.0&66.7 $\pm$47.1&\cellcolor{green!60}100.0$\pm$0.0&\cellcolor{green!15}80.0$\pm$40.0&60.0$\pm$49.0&56.7$\pm$49.6\\
				MM-BD(ours)&0& \cellcolor{green!60} 100.0$\pm$0.0& \cellcolor{green!60} 100.0$\pm$0.0& \cellcolor{green!60} 100.0$\pm$0.0& \cellcolor{green!60} 90.0$\pm$30.0& \cellcolor{green!60} 100.0$\pm$0.0& \cellcolor{green!60} 90.0$\pm$30.0& \cellcolor{green!60} 90.0$\pm$30.0&73.3$\pm$44.2& \cellcolor{green!60} 100.0$\pm$0.0& \cellcolor{green!15} 86.7$\pm$34.0& \cellcolor{green!60} 100.0$\pm$0.0\\
				\bottomrule		
		\end{tabular}
  \vspace{-0.1in}
	\caption{Detection accuracies (\%). For CIFAR-10, CIFAR-100, and GTSRB, 30 clean models and 30 models for each attack are trained. For TinyImageNet 10 clean models and 10 attack models are trained. Accuracy on clean models equals one minus the false positive rate.\vspace{-0.2in}}
		\label{tab:detection_accuracy}
	\end{center}
\end{table*}

In Tab. \ref{tab:detection_accuracy}, we show detection accuracy of MM-BD compared with the other methods on the backdoor attack ensembles we created and also report the number of legitimate images per class, $N_{\rm img}$, used by each method. A successful detection requires both the backdoor attack to be detected and the target class to be correctly inferred. We also show the proportion of clean classifiers deemed to be not attacked by each detector. We only evaluated META on CIFAR-10 since the computational cost for applying META to backdoor attacks on the other datasets is overly high (over 24 hours). For PT-RED, which estimates a backdoor pattern for each (source, target) class pair, we reduced its complexity by estimating a backdoor pattern only for each putative target class on CIFAR-100, TinyImageNet, and GTSRB -- otherwise, repeated experiments on these datasets will be infeasibly complex due to the large number of classes. Also due to the time constraint (and space limitations), we include detection results for only a few arbitrarily selected backdoor attack ensembles for CIFAR-100, TinyImageNet, and GTSRB, respectively. Complete results for MM-BD for {\it all} backdoor attack ensembles are in Apdx. \ref{sec:complete_results}.

As we have discussed in Sec. \ref{sec:related_work}, existing post-training detectors presume one or several backdoor pattern types. For example, NC shows strong capability in detecting backdoor attacks with patch replacement backdoor patterns (BadNet\&unicolor), for which it is designed; but NC fails to detect backdoor attacks with local additive backdoor patterns (1-pixel)\footnote{NC does detect global additive backdoor patterns when there are multiple source classes (similar results are reported in \cite{Post-TNNLS}).}. Similar results are reported for ABS and META, which are designed for patch replacement/blending backdoor patterns (BadNet, unicolor, and blend), while not addressing additive backdoor patterns (chessboard\&1-pixel). In contrast, PT-RED performs well for additive backdoor patterns (chessboard\&1-pixel), for which it is designed, but is generally ineffective for the other backdoor patterns (BadNet-blend). TABOR and TND do not show competitive performance compared with the other methods, since they adopt additional constraints on the shape or color of the backdoor pattern. Different from all these methods, MM-BD achieves {\it high detection accuracy} for all backdoor patterns with a {\it low false detection rate} for all datasets; {\it i.e.}, its performance is largely {\bf invariant to the backdoor pattern type}. Even if ABS (effective for patch backdoor patterns BadNet, unicolor and blend) and PT-RED (effective for additive backdoor patterns chessboard-1-pixel) are jointly deployed (detecting if either one detects), the detection accuracy on classifiers with backdoor attacks is just comparable to MM-BD, but with more false detections and a significant increment in computational cost. Moreover, NC, TABOR, ABS, and TND assume backdoor attacks have multiple source classes (the `M' setting); thus they may easily fail for backdoor attacks with a single source class (the `S' setting). However, MM-BD is generally effective in detecting backdoor attacks with {\bf arbitrary number of source classes}, as shown in Tab. \ref{tab:detection_accuracy}, as it does not make any assumptions about the number of source classes. Finally, different from all the other methods, MM-BD {\bf does not need any clean images for detection}.


\begin{table}[!t]
\vspace{0.1in}
	\begin{center}	\resizebox{\textwidth}{!}{
	\begin{tabular}{ccccc}
			\toprule
	&CIFAR10&CIFAR100&TinyImageNet&GTSRB\\
			\midrule
			NC &308.4$\pm$50.1s&800.4$\pm$147.5s&11227.0$\pm$1537.8s&412.5$\pm$51.4s\\
	TABOR&57.7$\pm$4.3s&341.1$\pm$25.2s&10792.2$\pm$824.6s&138.7$\pm$6.5s\\	ABS&50.3$\pm$3.2s&234.6$\pm$14.5s &819.1$\pm$91.7s&92.8$\pm$7.1s\\
	META&32h&- &-&-\\
	TND&591.9$\pm$16.6s&8207.3$\pm$257.2s&53530.8$\pm$1035.1s&1161.0$\pm$33.8s\\
	PT-RED&342.5$\pm$37.2s&-&-&-\\
	MM-BD (ours)&{\bf 27.2$\pm$3.4s}&{\bf 114.8$\pm$10.3s}&{\bf 503.4$\pm$42.1s}&{\bf 37.1$\pm$4.2s}\\
		\bottomrule
		\end{tabular}}
	\caption{Average execution times (with standard deviation).\vspace{-0.2in}} 
\label{tab:detection_efficiency}
	\end{center}
 \vspace{-0.1in}
\end{table}

In Tab. \ref{tab:detection_efficiency}, we show average execution times (in seconds). All experiments were conducted on a NVIDIA RTX-3090 GPU. Clearly, MM-BD is more efficient than most other post-training detectors. STRIP \cite{STRIP} is also known for its high efficiency, but it is an inference-time method, which detects if a test sample is embedded with a backdoor trigger, while our method detects if the {\it model} was attacked.
The inference time for META\cite{META} to inspect a single model is almost negligible. But for the unsupervised detection setting, in practice, the model to be inspected is typically associated with a new domain -- thus, META will need to train a large number of shadow models as well as the binary meta-classifier. This is very time-consuming.



\begin{table*}[!t]
\vspace{-0.2in}
	\fontsize{7pt}{7pt}\selectfont
        \centering
        \setlength{\tabcolsep}{2.5pt}
	\begin{tabular}{ccccccccccccc}
				\toprule
				
				&$N_{\rm img}$ &  & chessboard-S  & chessboard-M & 1-pixel-S & 1-pixel-M & BadNet-S &BadNet-M &unicolor-S&unicolor-M&blend-S&blend-M
				\\
				\midrule
				\multirow{2}{5.5em}{Without Mitigation}&\multirow{2}{0.5em}{} &ASR&99.9$\pm$0.1&99.9$\pm$0.1&91.1$\pm$7.8&91.3$\pm$6.9&99.4$\pm$0.3&99.9$\pm$0.1&97.8$\pm$1.3&98.4$\pm$1.6&96.4$\pm$2.8&96.9$\pm$3.4\\	&&ACC&91.3$\pm$0.9&91.1$\pm$0.8&91.8$\pm$0.4&91.1$\pm$0.3&91.6$\pm$0.4&91.6$\pm$0.3 &91.3$\pm$0.6&91.4$\pm$0.7&91.4$\pm$0.4&91.5$\pm$0.4\\
			\cline{3-13}\noalign{\vspace{0.5ex}}
				\multirow{2}{5.5em}{NC-M\cite{NC}}&\multirow{2}{.5em}{500}&ASR& 39.5$\pm$37.9&38.5$\pm$27.6&26.9$\pm$14.4&61.2$\pm$25.0&21.8$\pm$15.3&28.3$\pm$19.3&55.9$\pm$33.3&93.2$\pm$6.4&13.4$\pm$12.5&47.5$\pm$43.2\\
				&&ACC&87.0$\pm$1.9&87.7$\pm$0.9&90.8$\pm$0.5&86.0$\pm$4.2&84.9$\pm$4.9&76.7$\pm$6.5&86.0$\pm$1.1&85.1$\pm$5.4&88.4$\pm$2.1&88.2$\pm$1.6\\
				\cline{3-13}\noalign{\vspace{0.5ex}}
				\multirow{2}{5.5em}{Fine-Pruning\cite{FP}}&\multirow{2}{.5em}{500} &ASR&31.9$\pm$44.1&52.4$\pm$45.2&61.1$\pm$36.2&71.6$\pm$33.8&86.7$\pm$21.9&89.5$\pm$17.2&89.2$\pm$11.6&86.9$\pm$13.9&65.4$\pm$32.3&75.5$\pm$27.9\\
				&&ACC&90.7$\pm$1.0&90.6$\pm$0.7&91.2$\pm$1.0&91.5$\pm$0.6&91.2$\pm$1.0&91.6$\pm$0.5&91.3$\pm$0.2&90.9$\pm$1.8&91.6$\pm$0.7&91.6$\pm$0.6\\
                \cline{3-13}\noalign{\vspace{0.5ex}}
                \multirow{2}{5.5em}{NAD}&\multirow{2}{.5em}{250}& ASR& 3.2$\pm$2.5 & 3.0$\pm$2.0& 3.6$\pm$4.1&2.2$\pm$1.6 &{\bf 8.8}$\pm$6.7& {\bf 4.0}$\pm$3.6 &3.9$\pm$3.2 & 1.8$\pm$0.3 &{\bf 1.7} $\pm$0.7& {\bf 2.1}$\pm$1.8\\
				&&ACC&82.7$\pm$2.4 &81.1$\pm$2.5& 83.5$\pm$3.0&84.3$\pm$4.3& {\bf 87.5}$\pm$1.2&{\bf 88.1}$\pm$1.8  &82.5$\pm$3.1& 82.8$\pm$2.5&{\bf 87.0}$\pm$1.1& {\bf 88.4}$\pm$1.6\\
				\cline{3-13}\noalign{\vspace{0.5ex}}
				\multirow{2}{5.5em}{MM-BM(ours)}&\multirow{2}{.5em}{20} &ASR&99.4$\pm$0.3&99.5$\pm$0.2&{\bf 7.8}$\pm$4.9&{\bf 9.0}$\pm$3.2&{\bf 3.1}$\pm$5.0 &{\bf 1.8}$\pm$0.9&12.2$\pm$16.9&{\bf 5.0}$\pm$8.6&10.8$\pm$9.2&{\bf 9.9}$\pm$8.2\\
				&&ACC&90.4$\pm$1.5&90.7$\pm$1.4&{\bf 87.4}$\pm$1.2 &{\bf 88.3}$\pm$1.4 &{\bf 87.2}$\pm$2.8 &{\bf 90.6}$\pm$1.1 &89.8$\pm$0.4 &{\bf 89.8}$\pm$3.2&88.7$\pm$1.2 & {\bf 90.8}$\pm$1.1\\
				\cline{3-13}\noalign{\vspace{0.5ex}}
				\multirow{2}{5.5em}{MM-BM(ours)\\+\mbox{Fine-Pruning}}&\multirow{2}{.5em}{500}&ASR&55.2$\pm$40.1&53.1$\pm$36.3&{\bf 2.4}$\pm$2.8&{\bf 2.4}$\pm$4.2&{\bf 1.2}$\pm$0.5&{\bf 1.7}$\pm$1.0 &{\bf 1.2}$\pm$2.6&{\bf 1.2}$\pm$0.8&{\bf 2.3}$\pm$0.7& {\bf 2.3}$\pm$3.6\\
				&&ACC&90.1$\pm$0.7&90.2$\pm$0.7&{\bf 90.1}$\pm$0.6&{\bf 89.7}$\pm$0.9 &{\bf 90.0}$\pm$2.4 &{\bf 90.2}$\pm$0.8&{\bf 90.5}$\pm$0.6 &{\bf 90.2}$\pm$0.6&{\bf 89.6}$\pm$0.5 & {\bf 89.9}$\pm$1.2\\
				\bottomrule
		\end{tabular}	
  \caption{Average ASR(\%) and average ACC(\%) of classifiers for each of the ten backdoor attack ensembles created for CIFAR-10, after each of NC-M, FP, and MM-BM is applied. Results for which (jointly) ACC$\geq$87\% and  ASR$\leq$10\% are in bold. (All the test examples from the source classes with the backdoor trigger incorporated are used to measure the ASR).\vspace{-0.2in}} 
		\label{tab:mitigation}
\vspace{-0.1in}
\end{table*}

\subsection{Experiments for Backdoor Attack Mitigation}\label{subsec:exp_mitigation}

We evaluate our backdoor mitigation approach, dubbed ``MM-BM'' in the following, using the ten backdoor attack ensembles created in Sec. \ref{subsubsec:settings} for CIFAR-10, compared with 3 other backdoor mitigation approaches: NC-M (the mitigation approach associated with NC) \cite{NC}, fine-pruning (FP) \cite{FP}, and NAD \cite{NAD} (Notably, backdoor detection and mitigation are two different tasks; here we only consider mitigation methods). NC-M embeds the backdoor pattern reverse-engineered for the target class detected by NC into clean images from all source classes, and then fine-tunes the classifier deemed to be attacked on these images to ``unlearn'' the backdoor mapping. Here, we evaluate NC-M regardless of whether the backdoor attack is successfully detected by NC -- we apply NC-M to each backdoor attack in each ensemble using the backdoor pattern estimated by NC for the ground-truth target class. FP, on the other hand, removes neurons with low activations on clean images (these neurons are hypothesized to be ``reserved'' for backdoor patterns) subject to a prescribed budget of accuracy degradation; and then fine-tunes the classifier to recover its accuracy. Since FP does not perform backdoor attack detection, we directly apply it to all backdoor attack ensembles on CIFAR-10. NAD \cite{NAD} erases the backdoor attack by knowledge distillation. A {\it teacher} is first obtained by fine-tuning the attacked model on a small clean data set; then the attacked model is fine-tuned under the guidance of the teacher model's activations.
MM-BM is implemented following the description in Sec. \ref{subsec:backdoor_mitigation}, with the accuracy constraint set to $\pi=0.95$ and 20 clean images per class used for mitigation. Again, these settings are not critical to the performance of MM-BM. For the convolutional neural networks used in our experiments, we apply a common activation upper bound to all neurons associated with the same convolutional filter, since all the neuron activations in the feature map produced by a given convolutional filter should have the same dynamic range. More implementation details are deferred to Apdx. \ref{sec:details_MMBM} due to space limitations.

\begin{table}[!t]
\vspace{0.1in}
	\begin{center}	\resizebox{0.9\textwidth}{!}{
	\begin{tabular}{ccccc}
			\toprule
	&NC-M&fine-pruning&NAD&MM-BM(ours)\\
			\midrule
			time (s) &40.7$\pm$0.5&42.8$\pm$0.2&34.21$\pm$2.5&25.3$\pm$0.7 \\
		\bottomrule
		\end{tabular}}
\caption{Average execution times of MM-BD compared with other mitigation approaches on CIFAR-10.\vspace{-0.1in}} 
\label{tab:mitigation_efficiency}
	\end{center}
 \vspace{-0.05in}
\end{table}


\begin{figure}[!ht]
\includegraphics[width=.5\linewidth]{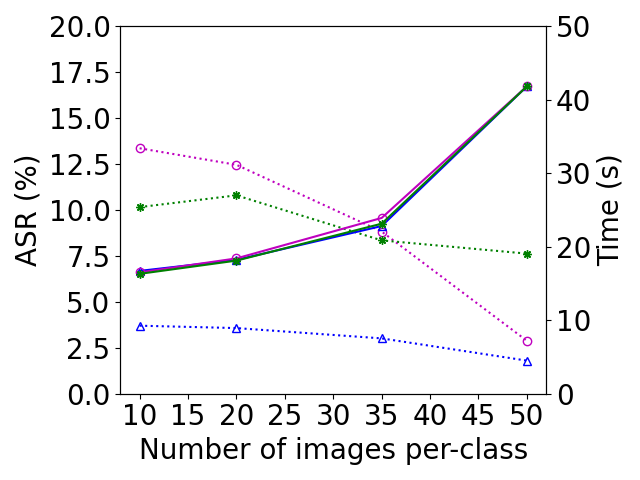} %
 \includegraphics[width=0.26\linewidth]{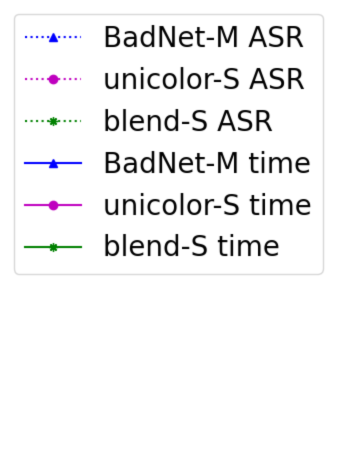}
	\caption{ASR and execution time for MM-BM versus the number of clean images per class used for mitigation, for three backdoor attack ensembles.\vspace{-0.25in}}%
\label{fig:mitigate_image_per_class}%
\vspace{-0.1in}
\end{figure}

In Tab. \ref{tab:mitigation}, we show the average ASR and average ACC of classifiers over each of the ten backdoor attack ensembles created on CIFAR-10, and after applying each of NC-M, FP, and MM-BM. We also show the number of clean images per class used by each method. Compared with the baseline without backdoor mitigation, NC-M significantly reduces ASR for most backdoor attack ensembles, although while suffering non-negligible degradation in ACC for some ensembles (notably for BadNet-M). In fact, the performance of NC-M largely relies on the effectiveness of the NC detector in backdoor pattern reverse-engineering and the hyper-parameter choices such as the step size for (un)learning. FP does not perform well in removing the backdoor mapping, possibly because the FP hypothesis that a subset of neurons are reserved solely for triggering the backdoor mapping does not hold. However, FP yields the highest ACC among the three mitigation approaches. The NAD method is generally effective for all types of backdoor patterns considered here (while MM-BM is not effective on chessboard-M and chessboard-S). However, the fine-tuning step of NAD (on a small clean dataset) causes large degradation in clean test accuracy. In comparison, with only 20 images per class (much fewer than for the other two methods), MM-BM reduces ASR to low levels for backdoor patterns 1-pixel-blend, with only moderate degradation in ACC. In Sec. \ref{subsubsec:design_choice_mitigation}, we will show that even better performance can be achieved by MM-BM for these backdoor patterns when a few more clean images are used. Unfortunately, MM-BM fails to mitigate backdoor attacks with the chessboard pattern (Notably, our detection method detects the chessboard pattern well, as shown in Table \ref{tab:detection_accuracy}). However, as shown in Apdx. \ref{sec:details_MMBM} and Fig. \ref{fig:two_stem_global} the internal activations caused by the chessboard pattern are just slightly larger than those induced by clean samples; thus, MM-BM, which places upper bounds on the activations of internal layers, cannot mitigate such a backdoor. But, given that MM-BM {\it does not} change the architecture or any trained parameters of the classifier, it can be naturally deployed together with other tuning-based mitigation methods. For example, as shown in Tab. \ref{tab:mitigation}, if MM-BM is deployed followed by FP, the ASR of all backdoor attacks will be largely reduced (and close to zero for backdoor patterns 1-pixel-blend), and with generally less degradation in ACC.

\vspace{-0.1in}
\subsubsection{Mitigation Design Choices}\label{subsubsec:design_choice_mitigation}

In Fig. \ref{fig:mitigate_image_per_class}, for ensembles BadNet-M, unicolor-S, and blend-S, we show ASRs and execution times versus the number of clean images used by MM-BM for backdoor mitigation. For 50 images per class, we observe clear drops in ASR, especially for the unicolor-S ensemble, with some (acceptable) increments in execution time. Even with as few as 10 images per class, MM-BM achieves decently low ASRs for all three ensembles.

\vspace{-0.1in}
\subsection{Detecting Advanced Backdoor Attacks}\label{subsec:advanced_attacks}
\vspace{-0.1in}

Here we consider advanced backdoor attacks including a clean-label/label-consistent backdoor attack\cite{Clean_Label_BA}, an input aware (IA) backdoor attack\cite{nguyen2020inputaware}, WaNet\cite{nguyen2021wanet}, and a label-smoothed backdoor attack\cite{label_smoothed}.

\begin{table}[h]
        \scriptsize
	\begin{center}
        \setlength{\tabcolsep}{2.5pt}
			\begin{tabular}{ccccccccccccc}
				\toprule
			& \multicolumn{4}{c}{clean label}& \multicolumn{1}{c}{IA} &   \multicolumn{1}{c}{WaNet}&\multicolumn{3}{c}{label smooth}   \\ \cmidrule(lr){2-5} \cmidrule(lr){6--6}\cmidrule(lr){7-7}\cmidrule(lr){8-10}
		PR	(\%)	& 0.4&1.5&6 &25&10&10&0.16&0.4&1
				\\
				\midrule
				ACC & 89.34& 89.20&88.34&88.68&90.73&91.02&91.41&91.14&90.97\\
                    ASR& 11.27&65.13&88.26&94.13&99.42&96.45&81.50&97.03&99.93\\
MM-BD&1/10&6/10&8/10&10/10&8/10&10/10&1/1&1/1&1/1\\
                    MM-BM ACC& 86.77&86.88&86.71&86.06&87.21&86.71&90.10&89.88&89.74\\
MM-BM ASR&1.68&1.65&1.64&1.82&12.30&8.74&2.32&3.50&1.13\\
				\bottomrule	
		\end{tabular}
	\caption{MM-BD and MM-BM performance on advanced attacks. PR represents the poisoning rate, the first and second rows show the ACC and ASR after attack, the third row shows the detection performance of MM-BD, the last two rows show the ACC and ASR after mitigation.\vspace{-0.1in}}
    \label{tab:advanced}
	\end{center}
 \vspace{-0.1in}
\end{table}
Unlike classical backdoor poisoning, the clean-label backdoor attack \cite{Clean_Label_BA} embeds the backdoor pattern into {\it target} class samples, and {\it without} label flipping. To ensure the backdoor pattern, not the original class-discriminating features of the target class, is learned, \cite{Clean_Label_BA} proposed to ``destroy'' those discriminating features using adversarial perturbations.


In \cite{nguyen2020inputaware}, an input-aware(IA)/sample-specific backdoor pattern was proposed. The backdoor pattern for each sample is obtained using a generator given the sample as the input. The generator and the DNN classifier are trained jointly such that: 1) the accuracy of the classifier on clean inputs is maximized; 2) 
the backdoor pattern generated for each sample induces the image to be misclassified to the designated target class; and 3) the backdoor pattern generated for one sample does not induce other samples to be misclassified. Based on this design, existing methods such as NC -- which reverse-engineers a common backdoor pattern in the input domain that induces a group of samples to be misclassified -- easily fail \cite{nguyen2020inputaware}. 


In \cite{nguyen2021wanet}, the backdoor pattern is generated for each sample using a `WaNet', and is smooth and largely imperceptible to humans. Also in \cite{nguyen2021wanet}, a `noise' training mode is proposed to help the attack evade some backdoor detectors. In particular, images with the backdoor pattern plus additive noise will not be (mis)classified to the designated target class. Thus, to detect backdoor attacks with such warping-based backdoor patterns using a reverse-engineering-based defense, the backdoor pattern used by the attacker should be precisely estimated, which is almost impossible in practice.


A label-smoothed backdoor attack was proposed in \cite{label_smoothed}, with more than one backdoor pattern associated with each attack. A similar embedding function as in \cite{BadNet} and \cite{nguyen2021wanet} was used, though the label flipping probability for each sample ${\bf x}_i\in{\mathcal X}$ is now some $p_n({\bf x}_i)\in[0, 1]$ instead of one. The label flipping probability can be determined using a surrogate classifier trained on an independent dataset possessed by the attacker. The purpose here is to prevent a single backdoor pattern from causing a high ASR when it appears in a test instance. However, when multiple backdoor patterns used for poisoning are simultaneously embedded in a test instance, the test instance will be (mis)classified to the backdoor target class with high confidence, leading to a high ASR.

The evaluation of MM-BD is conducted on CIFAR-10. For IA and WaNet, following the authors' suggestion, we used the PreActResNet-18 architecture; the ResNet-18 architecture was used for the other attacks. For clean label and label smooth attacks, we create multiple groups of attacked models with different poisoning rates. As shown in Table \ref{tab:advanced}, for attacks with high ASR (higher than 80\%), MM-BD achieves detection accuracy higher than 8/10; also, for all attacks, MM-BM reduces ASR to a low value with moderate degradation in ACC. Again, such good performance is based on the fact that MM-BD does not perform backdoor reverse-engineering and does not make assumptions about the backdoor pattern type.

\subsection{Detecting Attacked Models Trained with Differential Privacy}

The ``landscape'' of the classifier’s function will largely be altered when the model is trained with differential privacy. While differentially private training (e.g. DPSGD \cite{DPSGD}) may reduce overfitting to the exact backdoor pattern (and the target class margin accordingly), sufficient overfitting to the key features of the backdoor pattern is still required to ensure the backdoor pattern overrides the original class-discriminating features when embedded in any input (so that the backdoor pattern will be recognized). Thus, the margin for the backdoor target class will still be large enough to be detected by MM-BD.

Specifically,
following the standard implementation of DPSGD (with noise scale 0.1, and maximum gradient norm 2.0 — a larger noise scale will cause a severe drop in ASR), we trained 5 models on CIFAR-10 with the BadNet-M attack, resulting in ACC 58.06, ASR 87.61. We observe that the margins for all classes are significantly reduced compared to those for models trained without differential privacy. However, MM-BD still detects all 5 attacks.

\vspace{-0.1in}
\subsection{Detection Performance on Other Domains}
\label{subsec:other_domain}
\vspace{-0.1in}
While we mainly focus on image classification tasks in this paper, backdoor attacks (and defenses) have been extended to other domains such as natural language processing \cite{language_backdoor, badnl}, speech recognition \cite{abdullah2021sok}, video \cite{li2022few}, and point clouds \cite{xiang2021PCBA}. Here, we show the effectiveness of MM-BD against backdoor attacks on speeches and point clouds for example.

\subsubsection{Speech Command Classification}
We evaluate MM-BD on a speech domain using 10 clean and 10 backdoor ensembles of M5 \cite{m5_net} models trained on the Speech Commands dataset\cite{speech_commands}. The dimension of the backdoor pattern is 40 (the dimension of the speech signal is 8000). The pattern is embedded using the BadNet\cite{BadNet} embedding function, with a poisoning rate of 10\%. As shown in Tab. \ref{tab:other_domain}, our method has good detection performance on this speech domain.

\subsubsection{Point Cloud Classification}

On the point cloud domain, we use the attack proposed in \cite{xiang2021PCBA} to evaluate our detection method. The backdoor pattern is a
small set of inserted points (e.g.) to mimic real objects, such as a ball carried by a pedestrian. 10 good and 10 attacked DGCNN \cite{dgcnn} models were trained. Tab. \ref{tab:other_domain} shows that the detection performance for this domain is not as good as for speech. As revealed in \cite{xiang2021PCBA}, {\it intrinsic backdoors} exist in models trained on point cloud datasets, which means that, for a clean model, there will be a universal pattern that can induce misclassifications when embedded into test examples.  This may explain why the detection results are not as strong as for other domains.

Gradient-based MM-BD may not perform well on categorical
datasets depending on how their features are mapped to 
continuous numerical ones (e.g., natural language). In these cases, alternative
search strategies could be used by MM-BD, e.g., based on genetic algorithms.

\vspace{-0.1in}
\subsection{Robustness Against Adaptive Attacks}
\vspace{-0.1in}
\subsubsection{Adaptive Attack via Minimization of the Maximum Margin Statistic}
\label{subsubsec:adaptive}

Here, we consider a strong adaptive attack where the attacker has full knowledge of MM-BD. Moreover, we allow the attacker to have full control of the training process in addition to its standard poisoning capability. The purpose is to evaluate MM-BD more thoroughly, under extreme conditions, to show its superior power of post-training backdoor detection.

Note that a successful detection will be made by MM-BD if the maximum margin statistic associated with the backdoor target class is abnormally large. This abnormality can be easily observed for most backdoor settings and for a large range of backdoor pattern types as shown by our main experiments. Thus, to defeat MM-BD, the strong, adaptive attacker can fine-tune the classifier's parameters on the same poisoned training set (to keep both a high ASR and a high ACC) while {\it minimizing} the maximum margin for the backdoor target class. To do so, the attacker minimizes the following training loss:
\begin{equation}\label{eq:problem_min}
\begin{split}
    \min_{\phi}\quad  & \beta_{\rm T} \times \frac{1}{|{\mathcal D}_{\rm T}|}\sum_{({\bf x}, y)\in{\mathcal D}_{\rm T}}{\mathcal L}_{\rm E}(f({\bf x}; \phi), y)  \\& + \beta_{\rm B} \times \frac{1}{|{\mathcal D}_{\rm B}|}\sum_{(\tilde{\bf x}, t)\in{\mathcal D}_{\rm B}}{\mathcal L}_{\rm E}(f(\tilde{\bf x}, \phi), t) \\& + \beta_{\rm M} \times {\mathcal L}_{\rm M}(t; \phi),
\end{split}
\end{equation}
where
\begin{equation}\label{eq:problem_max}
    {\mathcal L}_{\rm M}(t; \phi) = \max_{{\bf x}\in{\mathcal X}} [g_t({\bf x}; \phi) - \max_{k\in{\mathcal Y}\setminus t} g_{k}({\bf x}; \phi)]
\end{equation}
is the maximum margin for class $t\in{\mathcal Y}$. Slightly different from the notations used previously, here, for both the classifier $f:{\mathcal X}\rightarrow{\mathcal Y}$ and the logit function $g_c:{\mathcal X}\rightarrow{\mathbb R}$ for class $c\in{\mathcal Y}$, we explicitly include the model parameters denoted by $\phi$. ${\mathcal D}_T$ is the set of clean training samples, with multiplier $\beta_{\rm T}$. ${\mathcal D}_B$ is the set of backdoor poisoned training samples, with multiplier $\beta_{\rm B}$. $\beta_M$ is the multiplier on the maximum margin term. ${\mathcal L}_{\rm E}$ denotes the cross-entropy loss. ${\tilde{\bf x}}$ and $t\in {\mathcal Y}$ are the backdoor-poisoned sample and the backdoor target class, respectively.

\begin{table}[t!]
	\fontsize{7pt}{7pt}\selectfont
	\begin{center}
			\begin{tabular}{ccccc}
				\toprule
			& \multicolumn{2}{c}{speech}& \multicolumn{2}{c}{point cloud} \\ \cmidrule(lr){2-3} \cmidrule(lr){4-5}
    		&clean&attacked&clean&attacked
				\\
				\midrule
			ACC& 81.4$\pm$0.5&80.3$\pm$1.0 &91.0$\pm$0.3& 87.5$\pm$0.5\\
                ASR&0&99.4$\pm$0.2&0 &99.0$\pm$0.8\\
                MM-BD& 10/10&10/10&8/10&7/10\\
				\bottomrule	
		\end{tabular}
  \vspace{-0.15in}
	\caption{Detection results for speech and point cloud.\vspace{-0.2in}}
    \label{tab:other_domain}
\end{center}
\end{table}

\begin{figure}[!t]
\vspace{-0.1in}
	\centering
	\subfloat[\centering ASR and ACC for attacks with different $\beta_{\rm M}$. ]{{\includegraphics[width=4cm]{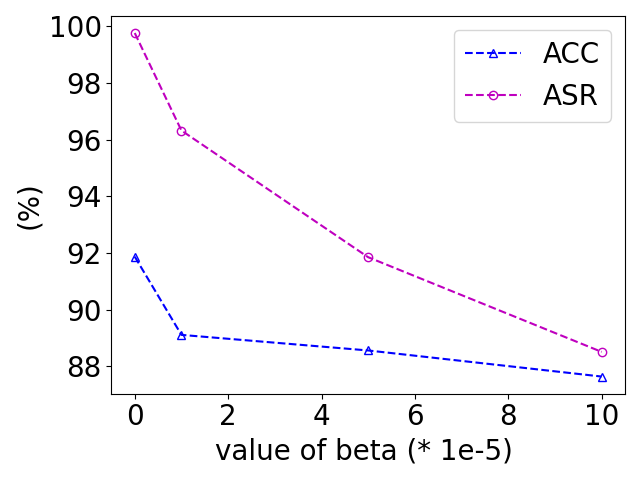} }\label{fig:adaptive_asr}}%
	~
	\subfloat[\centering maximum margin statistics for attacks with different $\beta_{\rm M}$. ]{{\includegraphics[width=4cm]{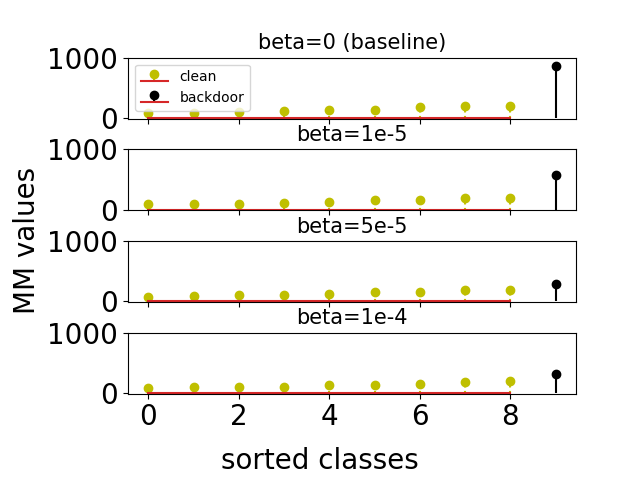} }\label{fig:adaptive_stem}}%
 \vspace{-0.1in}
	\caption{Results for adaptive attack.\vspace{-0.2in}}%
\end{figure}

To solve this min-max problem, the attacker can follow the classical strategy for adversarial training \cite{PGD} by maximizing Eq. (\ref{eq:problem_max}) with the model parameters $\phi$ fixed, and then minimizing Eq. (\ref{eq:problem_min}) to update the model. Note that the first two terms in Eq. (\ref{eq:problem_min}) are associated with traditional training on the backdoor poisoned training set to launch an attack. In our experiments, we consider $\beta_{\rm M} \in$ \{$10^{-5}$, $2\times 10^{-5}$, $10^{-4}$\}. As shown in Fig. \ref{fig:adaptive_asr}, as $\beta_{\rm M}$ increases, the abnormality of the maximum margin statistic associated with the backdoor target class decreases, such that attacks with $\beta_{\rm M}=10^{-4}$ are not detectable by MM-BD. However, there is a degradation in ASR of the classifier as $\beta_{\rm M}$ increases, with also some drop in the ACC.
There will be moderate drop in ASR and ACC for an adaptive attacker to bypass our detection method.
Moreover, the conventional backdoor 
simply requires
modifying a subset of the training data, while the adaptive attack requires the attacker to have full control of the training process.  
Also to defeat MM-BD, an attacker needs to solve the min-max optimization, which is time-consuming. Conventional training (60 epochs) takes 35 minutes while fine-tuning an attacked model based on the objective (\ref{eq:problem_min}) (for 1 epoch, which can make the attacked model undetectable) takes 62 minutes.
Finally, all these adaptive attacks are successfully mitigated by MM-BM, with average ACC of 85.8\% and average ASR of 2.3\% post-mitigation.

\subsubsection{A Stronger Adaptive Attack}
\label{subsubsec:stronger_adaptive}

The adaptive attack proposed above can be further enhanced by also maximizing the maximum margin of other classes. ${\mathcal L}_{\rm M}(t; \phi)$ in the last term of Eq. \ref{eq:problem_min} now becomes:
\begin{equation}\label{eq:enbanced_adaptive}
\begin{split}
    {\mathcal L}_{\rm M}(t; \phi)& = \max_{{\bf x}\in{\mathcal X}} [g_t({\bf x}; \phi) - \max_{k\in{\mathcal Y}\setminus t} g_{k}({\bf x}; \phi)] \\& - \sum_{s\in{\mathcal Y}\setminus t} \{\max_{{\bf x}\in{\mathcal X}} [g_s({\bf x}; \phi) - \max_{k\in{\mathcal Y}\setminus s} g_{k}({\bf x}; \phi)]\}.
\end{split}
\end{equation}

For $\beta_{\rm M} = 10^{-5}$, the learned model has ACC 88.75\% and ASR 96.04\% (with a similar drop in ACC but less drop in ASR than for the basic adaptive attack). This model is not detectable by MM-BD, with p-value 0.34.

While the
adaptive term defined in Eq. \ref{eq:enbanced_adaptive} leads to a stronger adaptive attack, it also makes the attack $|{\mathcal Y}|$ times more computationally expensive than the basic adaptive attack defined by Eq. \ref{eq:problem_max} and Eq. \ref{eq:problem_min} (since the attacker needs to perform maximum margin optimization for all the classes). It takes 642 minutes to fine-tune one epoch.

\vspace{-0.08in}
\section{Limitations and discussion}
\vspace{-0.08in}
\subsection{All-to-All Attacks}
\vspace{-0.08in}


For all-to-all attacks where every class is a backdoor target class with a specific\footnote{The very first all-to-all attack assumes that the backdoor target class for source class $i$ is class $(i+1){\rm mod}K$, where $K=|{\mathcal Y}|$ is the total number of classes. The scenario considered here is more general.} source class \cite{BadNet}, all the detection statistics are induced by the backdoor attack. So, there will be no statistics to provide any information about the null distribution for non-backdoor classes; unsupervised anomaly detection is clearly an ill-posed problem for this scenario.

\begin{figure}[!t]
\vspace{-0.2in}
	\centering
	\subfloat[\centering  ]{{\includegraphics[width=4.2cm]{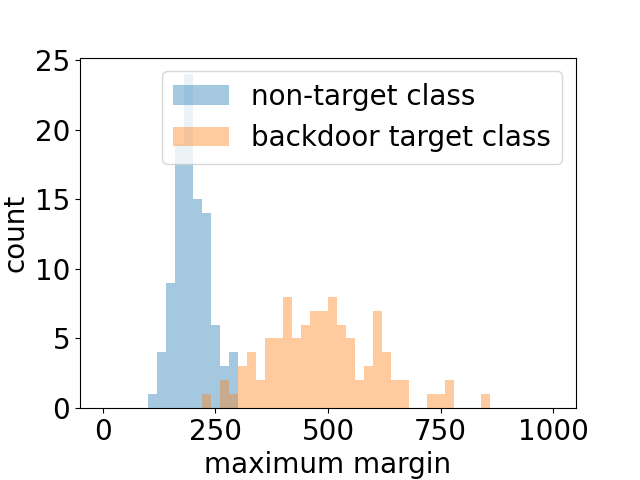} }\label{fig:all2all_diff}}%
	~
	\subfloat[\centering  ]{{\includegraphics[width=4.2cm]{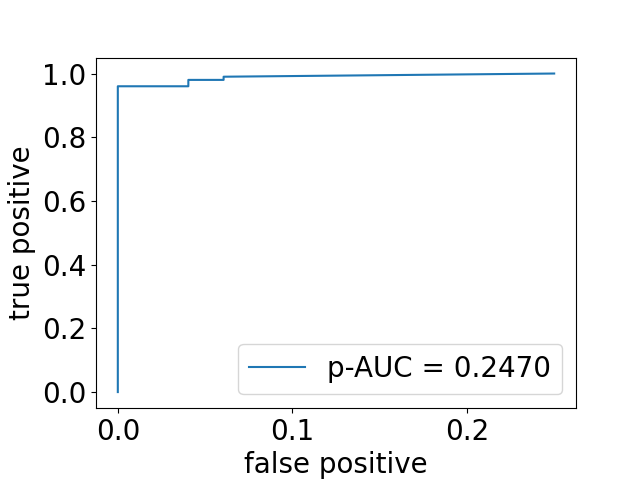} }\label{fig:roc_diff}}%
 \vspace{-0.1in}
	\caption{Analysis of maximum margin statistics for all-to-all attacks. (a) Histogram of maximum margin statistics for backdoor target classes and for non-target classes. (b) The ROC curve and the partial area under the curve (p-AUC) when the false positive rate is less than 0.25. \vspace{-0.1in}}
\end{figure}

Despite this, our maximum margin (MM) statistic can still be used for detection with a calibrated threshold leveraging some domain knowledge to set a detection threshold, {\it e.g.} as in \cite{TrojAI, Competition, Shen2021BackdoorSF, ABS}. Here, we created ten all-to-all attacks on CIFAR-10. For each attack, every class is a backdoor target class with a unique source class that is randomly assigned. In other words, each of the ten classes will be the source class for one backdoor class pair, and the target class for another. For each backdoor class pair, we randomly generated a backdoor pattern from type BadNet. In Fig. \ref{fig:all2all_diff}, we show the distribution of the maximum margin statistics obtained for each target class for each attack, compared with the maximum margin statistics obtained from ten clean classifiers. In Fig. \ref{fig:roc_diff}, we show the receiver operating characteristic (ROC) curve when the false positive rate is less than 0.25 in Fig. \ref{fig:all2all_diff}. The area under the curve (AUC) is 0.2470, showing that our MM statistic is very effective at distinguishing backdoor target classes from non-target classes regardless of the attack setting.

Moreover, we applied MM-BM to mitigate the ten attacks we created. Notably, MM-BM does not depend on the number of backdoor target classes by its design. It achieves 89.3\% average ACC and 1.7\% average ASR after mitigation -- all the all-to-all attacks are successfully mitigated.

\vspace{-0.1in}
\subsection{Potential False Detection: Case Studies}
\vspace{-0.1in}
MM-BD possesses neither clean models nor clean samples for adjusting the detection threshold. We use a detection threshold $\theta=0.05$ for all the datasets, which {\it theoretically} induces 5\% false positives. The same significance level is also adopted by NC, TABOR, and PT-RED. However, the theoretical false positive rates may not always be accurate due to inaccuracy of the estimated ``null'' model used for hypothesis testing. Especially, on CIFAR-10, there are only 9 MM statistics available for estimating the ``null'' model, with some false detections observed in the experiment shown in Table \ref{tab:detection_accuracy}. Notably, our method is generally better than the baseline methods in false positive control. Moreover, to achieve a lower false positive rate (FPR), one can always adopt a more conservative significance level to achieve a desired FPR, with only a moderate drop in the true positive rate (TPR). Tab. \ref{tab:fpr} gives the FPR and TPR over all models trained on CIFAR-10 under different significance levels.

On other datasets like CIFAR-100 and TinyImageNet, with more classes, more MM statistics are available for estimating the ``null'' model and false detections tend to be lower. However, GTSRB is an exception. Some false detections are still observed even though the number of classes in GTSRB is larger than  CIFAR-10 -- this could be attributable to class imbalance in GTSRB, which is discussed below. 

\begin{table}[t!]
	\scriptsize
	\begin{center}
		\begin{tabular}{ccccc}
			\toprule
                $\theta$ & 0.05 &0.02& 0.01 & 0.005 \\
                \midrule
                FPR (\%)&11.3  & 11.3&6.7&0\\
                TPR (\%)&91.3&88.0 &85.3& 82.3\\
			\bottomrule
		\end{tabular}
  \vspace{-0.1in}
		\caption{TPR and FPR under different significance levels.} 
		\label{tab:fpr}
	\end{center}
\end{table}

\subsubsection{Clean model trained on imbalanced data}
\label{subsubsec:class_imbalance}

\begin{figure}[!t]
	\centering
\includegraphics[width=.5\linewidth]{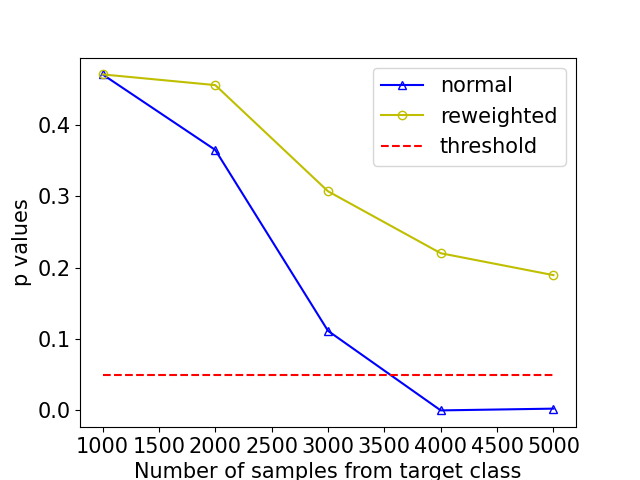} %
	\caption{Detection p-values of normally trained models (denoted as ``normal'' in the figure) and models trained with re-weighted data sampler (denoted as ``reweighted'') under different numbers of examples from the target class. The number of samples from classes other than the target class is fixed as 1000.\vspace{-0.25in}}%
	\label{fig:p_values}%
\end{figure}

In MM-BD, the abnormally large maximum margin is used as evidence of a backdoor attack. Notably, clean models trained on imbalanced data can also be overconfident on a particular class's decision, resulting in false detections by MM-BD. However, in practice, data re-weighted sampling is commonly applied to ``balance out'' imbalanced data. 
Here we evaluate the influence of class imbalance on MM-BD. We use a subset of CIFAR-10. First a ``target'' class is chosen and we make the number of samples from the ``target'' class larger than that from other classes. In all the experiments we set class ``9: truck'' as the ``target'' class, with the number of samples from all other classes fixed at 1000. We trained the model two ways -- both with or without data re-weighted sampling.
The detection p-values with different number of examples from the target class are given in Fig. \ref{fig:p_values}, showing that MM-BD is robust against a moderate level of class imbalance (which is also verified in the experiment on GTSRB shown in Tab. \ref{tab:detection_accuracy}.), but makes false detections when the class imbalance becomes severe. 

{\bf Notes:} Backdoor poisoning may also cause class imbalance. However, the abnormally larger MM statistic for the backdoor target class is due to the backdoor attack, not to class imbalance. To see this, we conduct an experiment where the poisoned training set is class-balanced (4000 samples per class). Our MM-BD achieves a p-value of $2.7 \times 10^{-7}$ against this attack, much smaller than the detection threshold.

\vspace{-0.1in}
\subsubsection{Regularized Clean Model with Abnormally Large Maximum Margin}

Other than class imbalance, a ``regularized'' clean model also can have an abnormally large maximum margin. To evaluate the influence of such regularization, we added a term to the (clean) training loss, such that the resulting model will have an abnormally large maximum margin for a certain class. The training objective is as follows:

\vspace{-0.2in}
\begin{equation}
\begin{split}
    \min_{\phi}\quad &\frac{1}{|{\mathcal D}_{\rm T}|}\sum_{({\bf x}, y)\in{\mathcal D}_{\rm T}}{\mathcal L}_{\rm E}(f({\bf x}; \phi), y) 
    \\&- \beta \times \max_{{\bf x}\in{\mathcal X}} [g_t({\bf x}; \phi) - \max_{k\in{\mathcal Y}\setminus t} g_{k}({\bf x}; \phi)],
\end{split}
\end{equation}
where the first term is the cross entropy loss, and the second term is the maximum margin of a predefined {\it target class} $t$. Class ``9'' was chosen to be the target class and $\beta$ was set to $10^{-5}$. The trained model has test accuracy of 91.42\% (similar to the clean unregularized models trained solely on the cross entropy loss), and we did observe an abnormally large maximum margin for the target class. The model is falsely detected, with a p-value of $1.2 \times 10^{-11}$. 
However, in practice, an honest training authority would not use this ``regularized'' objective function for training, unless this training authority is an insider seeking to bypass MM-BD.

\vspace{-0.1in}
\section{Conclusions}\label{sec:conclusions}
\vspace{-0.1in}

In this paper, we revealed a maximum margin statistic that substantially discriminates between clean and backdoor-attacked DNNs and, based upon which, we proposed a post-training backdoor detection method that makes no assumptions about the backdoor pattern type or method of incorporation. MM-BD is based on a novel maximum margin statistic that can be estimated without using any legitimate (clean) samples. It accurately and efficiently detects backdoor attacks with arbitrary numbers of source classes, as shown by our substantial experiments. Additionally, we proposed a backdoor mitigation approach which effectively removes most backdoor mappings. 
This method does leverage a small number of legitimate samples.

\vspace{-0.05in}
\section*{Ethics Statement}\label{ethics_statement}
\vspace{-0.1in}

The purpose of this research is to understand the behavior of deep learning systems facing malicious activities, and enhance their safety level. The backdoor attack considered in this paper is well-known, with open-sourced implementation code. Thus, publication of this paper will be beneficial to the community in defending against backdoor attacks. 

\vspace{-0.1in}
\section*{Acknowledgement}
\vspace{-0.1in}
This work was supported by NSF SBIR grant 2132294.
\vspace{-0.2in}
\bibliographystyle{IEEE}
\bibliography{ref}



\appendices





\vspace{-0.1in}
\begin{algorithm}[ht]

        \small
	\caption{Minimizing Eq. \eqref{eq:lagrangian_mitigation} for backdoor mitigation.}\label{alg:mitigation}
	\begin{algorithmic}[1]
		\State {\bf Inputs}: dataset ${\mathcal D}$, ``unbounded'' logit functions $\{g_c(\cdot)\}_{c\in{\mathcal Y}}$, accuracy constraint $\pi$, step size $\delta$, maximum iteration count $\tau_{\rm max}$, scaling factor $\alpha>1$.
		\State {\bf Initialization}: ${\bf Z}^{(0)}$ set large (e.g. 100), $\lambda^{(0)}$ set to a small positive number (e.g. $10^{-1}$), ${\bf Z}^{\ast}={\boldsymbol{\infty}}$.
		\For{$\tau = 0:\tau_{\rm max}-1$}
		\State ${\bf Z}^{(\tau+1)} = {\bf Z}^{(\tau)} - \delta \nabla_{\bf Z} L({\bf Z}^{(\tau)}, \lambda^{(\tau)}; {\mathcal D})$
		\If{$\frac{1}{|{\mathcal D}|}\sum_{({\bf x}, y)\in{\mathcal D}}{\mathds 1}[y=\argmax_{c\in{\mathcal Y}} \bar{g}_c({\bf x}; {\bf Z})] \geq \pi$}
		\State $\lambda^{(\tau+1)} = \lambda^{(\tau)}\cdot\alpha$
		\If{$\sum_{l=2}^L (||{\bf z}^{(0)}_l||_2 - ||{\bf z}^{\ast}_l)||_2<0$}
		\State ${\bf Z}^{\ast} = {\bf Z}^{(\tau+1)}$
		\EndIf
		\Else
		\State $\lambda^{(\tau+1)} = \lambda^{(\tau)}/\alpha$
		\EndIf
		\EndFor
		\State {\bf Outputs}: ${\bf Z}^{\ast}$ \vspace{-0.05in}
	\end{algorithmic}
\end{algorithm}

\section{Experiment Details}
\vspace{-0.1in}

\subsection{Details of Datasets}\label{sec:detail_dataset}
\vspace{-0.1in}
CIFAR-10 \cite{CIFAR10} contains 60000 $32\times 32$ color images from 10 classes. For each class, 5000 images are for training and 1000 images are for testing. Similar to CIFAR-10, CIFAR-100 contains 60000 $32\times 32$ color images, with each class containing 500 training images and 100 test images\cite{CIFAR10}. GTSRB\cite{GTSRB} is a dataset containing color images of German traffic signs from 43 classes. There are 39209 training images and 12630 test images. TinyImageNet is a subset of ImageNet\cite{ImageNet}, but with each image resized to $64 \times 64$ (for the official version); there are 200 classes in the dataset, each containing 500 training images and 25 test images.
\begin{figure}[ht]
	\centering
	\begin{minipage}[b]{.17\linewidth}
		\centering
		\centerline{\includegraphics[width=\linewidth]{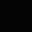}}
		\centerline{\includegraphics[width=\linewidth]{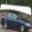}}
		\footnotesize{ ``chessboard''}\label{fig:bd_example_A1}
	\end{minipage}
	\begin{minipage}[b]{0.17\linewidth}
		\centering
		\centerline{\includegraphics[width=\linewidth]{Figures/bd_example/patternA2.png}}
		\centerline{\includegraphics[width=\linewidth]{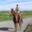}}
		\footnotesize{``1-pixel''}\label{fig:bd_example_A2}
	\end{minipage}
	\begin{minipage}[b]{0.17\linewidth}
		\centering
		\centerline{\includegraphics[width=\linewidth]{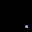}}
		\centerline{\includegraphics[width=\linewidth]{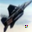}}
		\footnotesize{BadNet}\label{fig:bd_example_A3}
	\end{minipage}
	\begin{minipage}[b]{0.17\linewidth}
		\centering
		\centerline{\includegraphics[width=\linewidth]{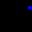}}
		\centerline{\includegraphics[width=\linewidth]{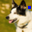}}
		\footnotesize{unicolor }\label{fig:bd_example_A4}
	\end{minipage}
	\begin{minipage}[b]{0.17\linewidth}
		\centering
		\centerline{\includegraphics[width=\linewidth]{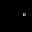}}
		\centerline{\includegraphics[width=\linewidth]{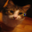}}
		\footnotesize{blended }\label{fig:bd_example_A5}
	\end{minipage}
	\caption{Example BPs used in our experiments (top) and images with these BPs embedded (bottom).\vspace{-0.1in}}
	\label{fig:bd_example}
\end{figure}
\vspace{-0.1in}
\subsection{Details of Backdoor Patterns}\label{sec:detail_bp}
\vspace{-0.1in}
chessboard and 1-pixel are two additive backdoor patterns. chessboard is a `chessboard' pattern considered in \cite{Post-TNNLS}, with one and only one of two
adjacent pixels perturbed positively by 3/255 in all color channels. 1-pixel is the single-pixel additive backdoor pattern used by \cite{SS}, \cite{AC} -- one pixel, randomly selected and fixed for all images for each attack, is perturbed positively by 75/255 in all color channels. BadNet and unicolor are both patch replacement backdoor patterns embedded by $\tilde{\bf x} = (1 - {\bf m})\odot{\bf x} + {\bf m}\odot{\bf u}$. Here, ${\bf u}$ is a random noise patch for BadNet and a monochromatic patch for unicolor. For BadNet, the patch size is $5\times5$ for the TinyImageNet dataset and $3 \times 3$ for other datasets. For unicolor, we use a $3 \times 3$ patch size for all datasets (though we did not apply unicolor to attack the TinyImageNet dataset due to the training cost). blend is a blended backdoor pattern used in \cite{META} and \cite{Targeted}. The blended mask is chosen to be a $3 \times 3$ square patch with a randomly selected and fixed location for each image in each attack. The blending rate is set to $\alpha = 0.2$ in our experiments. Examples of the BPs and attacked images of chessboard-blend are shown in Fig. \ref{fig:bd_example}.
\vspace{-0.1in}
\subsection{Details of Backdoor Attack configurations}
\label{sec:ba_config}
\vspace{-0.1in}
The number of backdoor training images used for poisoning was carefully chosen for each backdoor pattern and for each dataset to ensure a high attack success rate for the created backdoor attacks. Details are shown in Table \ref{tab:asr-acc}. 

%
%

\vspace{-0.1in}
\subsection{Training Configurations}\label{sec:training_config}
\vspace{-0.1in}
For each dataset, we use the same training configuration to train both clean and attack models. These details are shown in Table \ref{tab:train_config}.  

\begin{table}[!ht]
	\scriptsize
	\begin{center}
 \setlength{\tabcolsep}{2.5pt}
		\begin{tabular}{ccccc}
			\toprule
			&CIFAR-10&CIFAR100&TinyImageNet&GTSRB\\
			\midrule
			model & ResNet-18 & VGG-16 &ResNet-34 & MobileNet\\
			optimizer& Adam & Adam & Adam & Adam\\
			batch size&128 & 128 & 64 & 128\\
			epochs&60 & 100 & 90 & 50\\
			learning rate&1e-3 &1e-4 & 1e-3&1e-3\\
			\bottomrule
		\end{tabular}
  \vspace{-0.1in}
		\caption{Training configurations for the four datasets considered in our experiments.\vspace{-0.2in}} 
	\label{tab:train_config}
	\end{center}
\end{table}

\begin{table*}[h]
\vspace{-0.4in}
	\scriptsize
	\begin{center}	
 \setlength{\tabcolsep}{2.5pt}\begin{tabular}{cccccccccccc}
				\toprule
				&  \multicolumn{11}{c}{CIFAR-10}        \\ \cmidrule(lr){2-12} 
				& clean & chessboard-S  & chessboard-M & 1-pixel-S & 1-pixel-M & BadNet-S &BadNet-M &unicolor-S&unicolor-M&blend-S&blend-M
				\\
				\midrule
    poison \# &0&2000&2000&1000&1000&500&500&500&500&1000&1000\\
				ASR& 0 &99.94&99.94&91.09&91.28&99.41&99.92&97.78&98.44&96.36&96.86\\
				ACC&91.64&91.31&91.06&91.80&91.12&91.63&91.59 &91.31&91.36&91.35&91.48\\
				\bottomrule
                    \toprule				
				& \multicolumn{4}{c}{CIFAR-100}& \multicolumn{2}{c}{TinyImageNet} &   \multicolumn{5}{c}{GTSRB}   \\ \cmidrule(lr){2-5} \cmidrule(lr){6-7}\cmidrule(lr){8-12}
				&  clean & 1-pixel-M  & BadNet-M & unicolor-M & clean & BadNet-M &clean &chessboard-M&1-pixel-M&unicolor-M&blend-M
				\\
				\midrule
		poison \#&0&990&990&990&990&0&500&1680&840&840&840\\	ASR&0&95.67&99.84&97.37&0&97.84&0&99.80&99.20&95.80&100\\
				ACC&67.51&65.40&66.17&66.51&60.71& 58.59&94.78&94.54&94.12&94.56&94.53\\
				\bottomrule
		\end{tabular}
  \vspace{-0.1in}
		\caption{Average ASR and ACC over each ensemble for the four datasets.} 
		\label{tab:asr-acc}
	\end{center}
\end{table*}

\begin{table*}[h]
	\scriptsize
	\begin{center}
		\begin{tabular}{ccccccccccc}
			\toprule
			& &\multicolumn{3}{c}{CIFAR-100}& TinyImageNet& \multicolumn{4}{c}{GTSRB}   \\ \cmidrule(lr){3-5} \cmidrule(lr){6-6}\cmidrule(lr){7-10}
			
			& &1-pixel-M  & BadNet-M & unicolor-M  & BadNet-M  &chessboard-M&1-pixel-M&unicolor-M&blend-M
			\\
			\midrule
&ASR&95.7$\rightarrow$13.5&99.8$\rightarrow${\bf 1.5}&97.4$\rightarrow$16.9&97.8$\rightarrow${\bf 1.3}&99.8$\rightarrow$78.2&99.2$\rightarrow$26.8&95.8$\rightarrow$11.3&100$\rightarrow${\bf 1.5}\\
				&ACC&65.4$\rightarrow$61.7&66.2$\rightarrow${\bf 65.0}&66.5$\rightarrow$64.9& 58.59$\rightarrow${\bf 58.0}&94.5$\rightarrow$94.0&94.1$\rightarrow$95.0&94.6$\rightarrow$95.2&94.5$\rightarrow${\bf 95.4}\\
			
			\bottomrule
		\end{tabular}
		\caption{MM-BM results on the CIFAR-100, GRSTB, and TinyImageNet ensembles. Joint results with ACC drop $\leq$5\% and ASR$\leq$10\% are shown in bold.\vspace{-0.2in}} 
		\label{tab:complete_mitigation}
	\end{center}
 \vspace{-0.1in}
\end{table*}
\vspace{-0.1in}
\section{Additional Results}
\vspace{-0.1in}
\subsection{Complete Detection Results for MM-BD for all
	Backdoor Attack Ensembles}
\label{sec:complete_results}


Here, we show the complete detection results for MM-BD on all backdoor attack ensembles for CIFAR-10 and GTSRB in Tab. \ref{tab:complete_res}. MM-BD achieves generally high detection accuracy for most backdoor attack ensembles.

\begin{table}[h]
\vspace{0.15in}
	\scriptsize
	\begin{center}
 \setlength{\tabcolsep}{2.5pt}
		\begin{tabular}{ccccccc}
			\toprule
			&clean&  chessboard-M &1-pixel-M &BadNet-M &unicolor-M & blend-M 
			\\
			\hline 
			CIFAR-100&10/10&6/10& 10/10&10/10&10/10 & 10/10\\
                \hline
                GTSRB &17/20&7/10&10/10&10/10&9/10& 10/10\\
			\bottomrule
		\end{tabular}\vspace{-0.1in}
		\caption{Complete results of MM-BD on all backdoor attack ensembles created for CIFAR-100 and GTSRB.} 
		\label{tab:complete_res}
	\end{center}
\end{table}

\vspace{-0.05in}
\subsection{Details for MM-BM and Discussion}\label{sec:details_MMBM}
\begin{figure}[!ht]
\vspace{-0.15in}
	\centering
	\subfloat[\centering BadNet-M ]{{\includegraphics[width=4cm]{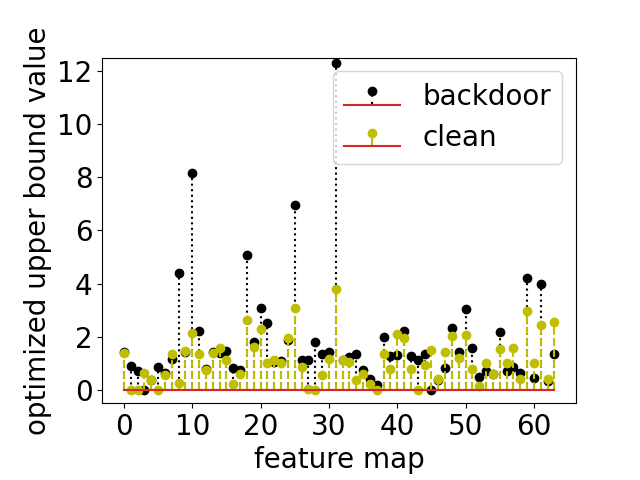} }\label{fig:two_stem_local}}%
	~
	\subfloat[\centering chessboard-M]{{\includegraphics[width=4cm]{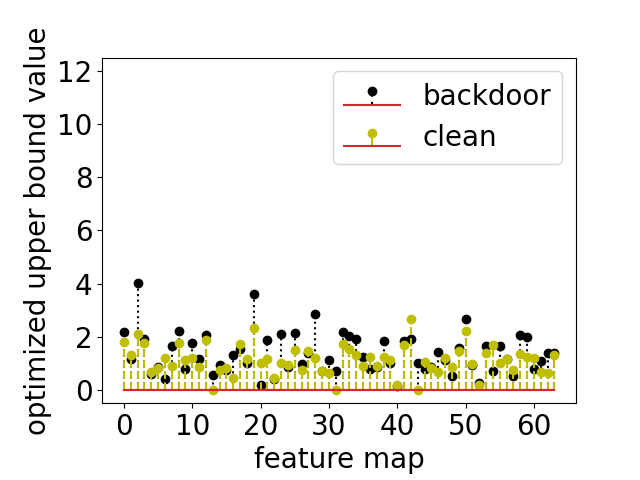} }\label{fig:two_stem_global}}%
	\caption{Stem plots for the optimized upper bounds using MM-BM with clean samples and backdoor poisoned samples for BadNet-M and chessboard-M respectively.\vspace{-0.2in}}%
 \vspace{-0.1in}
\end{figure}

\vspace{-0.15in}
In our experiments, we arbitrarily selected three layers close to the input to apply upper bounds on activations. For the experiments on CIFAR-10, for example, these three layers are the 1st, 5th, and 9th convolutional layers. In Fig. \ref{fig:two_stem_local}, we show a stem plot for the bounding vector optimized by MM-BM obtained for the first convolutional layer (marked as ``clean'') for an attack with the local backdoor pattern BadNet. In comparison, also in Fig. \ref{fig:two_stem_local}, we show a stem plot of the optimized bounding vector obtained from the same layer, but with the optimization problem Eq. (\ref{eq:opt_mitigation}) solved using samples embedded with the true backdoor pattern and labeled to the backdoor target class (marked as ``backdoor''). In this case, the optimized upper bounds represent the activation value required for a backdoor pattern to induce (mis)classification to the backdoor target class. Clearly from the figures, activations of a subset of neurons associated with the backdoor pattern are depressed by applying MM-BM with clean samples, which prevents the backdoor pattern from triggering misclassifications during testing. Thus, the attack is successfully mitigated.

Unfortunately, MM-BM does not effectively mitigate the chessboard-S and chessboard-M attacks. This is likely due to the per-pixel perturbation size of the global backdoor pattern (i.e. chessboard) being very tiny (merely 3/255). Such a tiny perturbation size will induce less significant changes to neuron activations compared with local patterns with large perturbation size (or patch replacement patterns that induces large modifications in pixel values). Thus, to ensure a high attack success rate, backdoor patterns like chessboard will be learned by the classifier to induce changes to the activations of a relatively large subset of neurons, where each neuron activation is only changed moderately. This hypothesis is well-supported by the stem plots in Fig. \ref{fig:two_stem_global}, where we repeated the process from Fig. \ref{fig:two_stem_local} for an attack with backdoor pattern chessboard. The differences in the two stem plots in Fig. \ref{fig:two_stem_global} are much smaller than for Fig. \ref{fig:two_stem_local}. Thus, when MM-BM is applied to this attack, the backdoor pattern will still be partially activated, leading to a poor mitigation result.

\vspace{-0.1in}
\subsection{Mitigation Performance of on other Datasets}\label{sec:mitigation_complete}
\vspace{-0.1in}

In Tab. \ref{tab:complete_mitigation}, we report the mitigation results of MM-BM for the other three datasets. Here, for CIFAR-100 and GTSRB, we apply activation upper bounds to the first three convolutional layers, since this will be more efficient than applying activation upper bounds to all layers, as discussed in Apdx. \ref{sec:details_MMBM}. Again, MM-BM is effective at mitigating backdoor attacks for most backdoor patterns.

\vspace{-0.05in}
\section{The hyper-parameters in the MM-BM method}
\vspace{-0.05in}
In our mitigation method (described in detail in algorithm \ref{alg:mitigation}, there are two hyper-parameters--$\lambda$ and $\alpha$. $\lambda$ is the weight on the $L_2$ norm of the upper bounds -- a larger $\lambda$ leads to lower upper bounds but worse accuracy following mitigation. $\alpha$ is a ``step size'' helping to adjust $\lambda$ ($\alpha$ needs to be larger than 1), which is the most important hyper-parameter in our MM-BM method.

\begin{figure}[!t]
\vspace{-0.05in}
	\centering
	\includegraphics[width=.5\linewidth]{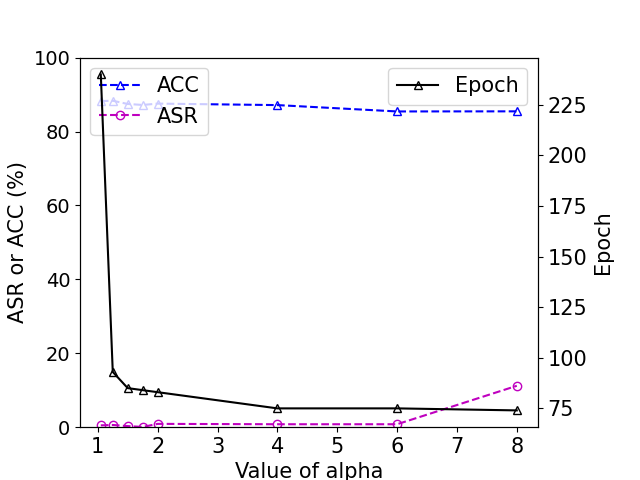} %
	\caption{MM-BD mitigation performance and the number of epochs before the stopping condition is reached, under different $\alpha$. (The attacked model's ACC is 91.25\% and ASR is 99.98\% before mitigation.)\vspace{-0.15in}}%
	\label{fig:alpha}%
 \vspace{-0.05in}
\end{figure}

Here we did an experiment to evaluate the influence of $\alpha$ on our mitigation method. Fig. \ref{fig:alpha} shows the ACC, ASR after mitigation and the number of epochs required to reach the stopping condition under different $\alpha$.
The results indicate that a larger $\alpha$ makes the mitigation converge more quickly, but a too large $\alpha$ might lead to a drop in performance. The results also show that a wide range of $\alpha$ values can lead to good mitigation results and high efficiency. Moreover, our mitigation method is not time-consuming, so in practice, when a validation set is not available, it is safe to use a small value of $\alpha$ to guarantee good performance.

\begin{figure}[!t]
	\centering
	\subfloat[\centering  ]{{\includegraphics[width=4cm]{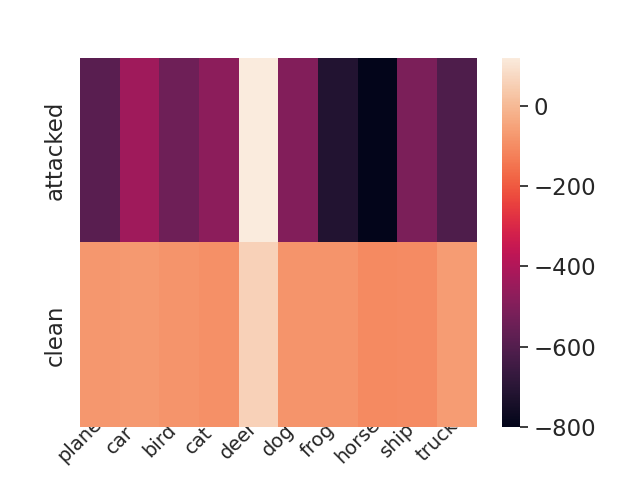} }\label{fig:logit_boosting_suppression}}%
	~
	\subfloat[\centering ]{{\includegraphics[width=4cm]{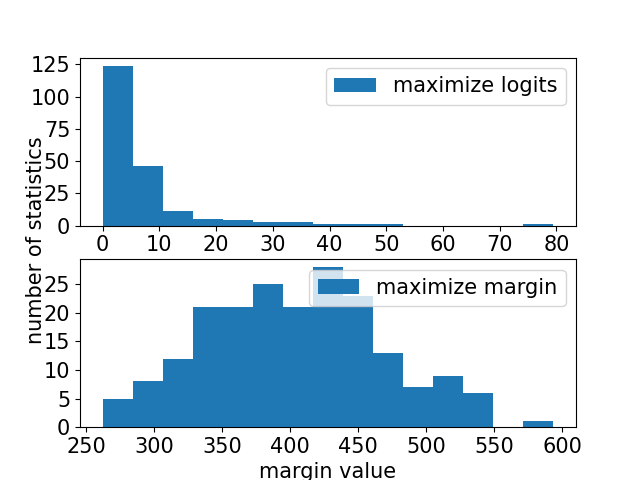} }\label{fig:hist_margin}}%
	\caption{(a): Logits of a classifier from 1-pixel-S with target class `deer' (top) and a clean classifier (bottom), when the logit of `deer' class is maximized. (b): Histogram of margin statistics of a clean classifier trained on TinyImageNet using different objective functions.
 \vspace{-0.2in}}%
 \vspace{-0.1in}
\end{figure}

\section{Empirical Analysis of Maximum Margin}\label{subsubsec:ablation}
\vspace{-0.1in}

Here, we first show that the existence of a backdoor attack boosts the target class logit, while more importantly, suppressing the logits of all other classes. We consider a classifier attacked with target class ``deer'' from the 1-pixel-S ensemble and a clean classifier, both for the CIFAR-10 domain. For both classifiers, we maximize the logit of the ``deer'' class using the same MM-BD protocol.  The resulting logits for all classes for the two classifiers are shown in Fig. \ref{fig:logit_boosting_suppression}. Even though we do not deliberately suppress the logits of all classes other than the ``deer'' class, we observe significant decrements in these logits compared with the clean case. Thus, while a backdoor attack may evade a detector that is based solely on maximizing the logit (when the ``boosting'' effect alone does not generate a sufficiently atypical statistic that is detectable), it will be easily detected by MM-BD based on MM, which leverages both logit ``boosting'' and ``suppression'' effects.

However, can we maximize just the logit and then obtain a margin statistic for detection? Unfortunately, this alternative will easily lead to false detections, especially when a substantial number of classes have closely ``neighboring'' classes containing similar semantic features. Here, we consider a clean classifier trained on TinyImageNet for example. By maximizing just the logit (i.e., maximizing the first term of \eqref{eq:opt_main}) for, e.g., the `plunger' class, we will obtain a similarly large logit for the `drumstick' class, possibly because the two categories of objects are similar to each other. Thus, the margin statistics obtained for the `plunger' class will be small. Given a large number of such small margins (e.g. the top of Fig. \ref{fig:hist_margin}), relatively large margins produced by classes without a ``neighboring'' class will likely appear as outliers, which will easily lead to false detections. By contrast, MM-BD avoids false detections by directly maximizing the margin instead of the logit, which yields decently large margins for most classes, as shown in the bottom of Fig. \ref{fig:hist_margin}.

\section{Meta-Review}

\subsection{Summary}
The paper describes methods to counter backdoors in already trained and distributed machine learning models, based on "maximum margin" statistics, that is, statistics across the classifier's logits. The authors observe that the target class of a backdoor model yields higher outputs than all other classes if maximized with gradient ascent and random initialization.

\subsection{Scientific Contributions}
\begin{itemize}
\item Addresses a Long-Known Issue
\end{itemize}

\subsection{Reasons for Acceptance}
\begin{enumerate}
\item {\bf Simplicity of the approach.} The underlying rationale seems apparent and intuitively plausible. MM-BD merely needs comparing logits which is beneficial for the approach's runtime complexity. Moreover, the defender neither knows the training data nor has details of the training procedure.
\item {\bf Extensive number of experiments.} The paper extensively evaluates the approach in various settings with different backdoor and trigger types. The authors extend to conduct adaptive attacks to explore the limits of their detection scheme.
\item {\bf Outperforms current state of the art.}
\end{enumerate}

\subsection{Noteworthy Concerns} 
\begin{enumerate} 
\item {\bf Not robust against adaptive attacks.} The presented results show that MM-BD is severely impacted by an adaptive attacker controlling the learning process.
\item {\bf High false positive rate for datasets with few classes.} The method exhibits a relatively large false-positive rate for datasets with a small number of class, which however is not the case for datasets with many classes.
\item {\bf Existing concurrent work:} Chen et al., "Effective Backdoor Defense by Exploiting Sensitivity of Poisoned Samples", NeurIPS 2022
\end{enumerate}

\section{Response to the Meta-Review} 

We explicitly accept the meta-review in its current form.

\end{document}